\documentclass[12pt]{spieman}  % 12pt font required by SPIE;
\usepackage{cite}
\usepackage{amsmath,amssymb,amsfonts}
\usepackage{algorithmic}
\usepackage{graphicx}
\usepackage{textcomp}
\usepackage{xcolor}
\usepackage{adjustbox} % for shrink-only sizing
\usepackage{ragged2e}
\usepackage{siunitx}
\usepackage{array}
\usepackage{multirow}
\usepackage{caption}
\usepackage{booktabs,tabularx,makecell}
\newcolumntype{Y}{>{\raggedright\arraybackslash}X}
\def\BibTeX{{\rm B\kern-.05em{\sc i\kern-.025em b}\kern-.08em
    T\kern-.1667em\lower.7ex\hbox{E}\kern-.125emX}}

\newlength{\figepsilon}     % safety margin to avoid float-too-large warnings
\newlength{\imgmaxheight}   % computed per figure
\newcommand{\computeimgmaxheight}{%
  \setlength{\imgmaxheight}{%
    \dimexpr\textheight-\abovecaptionskip-\belowcaptionskip-\figepsilon\relax}%
}

\begin{document}

\title{Information-driven Fusion of Pathology Foundation Models for Enhanced Disease Characterization}

\author[a,*]{Brennan Flannery}
\author[a]{Thomas DeSilvio}
\author[b]{Jane Nguyen}
\author[a,c,d,e]{Satish E. Viswanath}

\affil[a]{Case Western Reserve University, Department of Biomedical Engineering, 10900 Euclid Ave, Cleveland, OH, United States of America}
\affil[b]{Cleveland Clinic, Department of Pathology, 9500 Euclid Ave, Cleveland, OH, United States of America}
\affil[c]{Emory University, Department of Pediatrics, Atlanta, GA, United States of America}
\affil[d]{Emory University, Department of Biomedical Engineering, Atlanta, GA, United States of America}
\affil[e]{Louis Stokes VA Cleveland Medical Center, Cleveland, OH, United States of America}

\maketitle

\begin{abstract}
Foundation models (FMs) for digital pathology have demonstrated strong performance across diverse tasks, with many models being developed in recent studies. While there are similarities in the pre-training objectives of the FMs across these studies, there is still a limited understanding of complementarity between FM representations, the potential redundancy in their embedding spaces, or biological interpretation of their features. In this study, we propose an information-driven, intelligent fusion strategy for integrating multiple pathology FM embeddings into a unified representation as well as a systematic evaluation of its performance for cancer grading and staging across three distinct diseases. Diagnostic hematoxylin and eosin whole-slide images from publicly available TCGA-KIRC (kidney; 519 slides, 242 patients), TCGA-PRAD (prostate; 490 slides, 490 patients), and TCGA-READ (rectal; 200 slides, 200 patients) were dichotomized into low versus high grade or stage. Both tile-level FMs (Conch v1.5, MUSK, Virchow2, H-Optimus1, Prov-Gigapath) as well as slide-level FMs (TITAN, CHIEF, MADELEINE) were considered to train downstream classifiers. We then evaluated three FM fusion schemes at both tile and slide levels: majority-vote ensembling, naive feature concatenation, and intelligent fusion based on correlation-guided pruning of redundant features. When evaluated via patient-stratified cross-validation with held-out testing, intelligent fusion of tile-level embeddings yielded consistent, statistically significant gains in F1 score and AUC across all three cancers compared with the best single FMs and naive fusion, despite retaining as little as 1\% of the original FM feature spaces in some disease contexts. Global similarity metrics further revealed substantial alignment of tile-level embedding spaces, contrasted by lower local neighborhood agreement, indicating complementary fine-grained information across FMs and explaining performance gains from intelligent fusion. Attention maps revealed that intelligent fusion of FMs resulted in concentrated attention on tumor regions while also reducing spurious focus on benign regions. Unsupervised clustering in the intelligently fused FM space also showed significantly improved separation of tumor vs benign tiles compared to any alternative strategy. Our findings suggest that intelligent, correlation-guided fusion of pathology FMs can yield compact, task-tailored representations that enhance both predictive performance and interpretability in downstream computational pathology tasks.
\end{abstract}


\section{Introduction}
Histopathologic assessment remains central to disease evaluation, wherein tissue sections are stained, mounted on glass, and examined via light microscopy. Routine digitization of these sections yields whole-slide images (WSIs) at micrometer resolution that capture salient cellular and architectural phenotypes linked to tumor aggressiveness, stage, and other clinically relevant attributes ~\cite{cui_artificial_2021}. In current practice, expert pathologists synthesize these cues into narrative reports; however, the process is time-intensive~\cite{trpkov_novel_2021} and subject to inter-observer variability~\cite{kweldam_gleason_2016, ozkan_interobserver_2016}, particularly in borderline or heterogeneous lesions. Computational pathology has therefore emerged as a complementary strategy to standardize and accelerate interpretation by mining quantitative descriptors directly from WSIs ~\cite{cui_artificial_2021}.

The scale and resolution of WSIs impose nontrivial computational constraints. A single slide often comprises gigapixels of data, precluding naïve end-to-end processing on contemporary graphical processing units for whole slide-level predictions ~\cite{litjens_decade_2022}. To contend with these limitations, many pipelines decompose slides into fixed-size, non-overlapping image tiles that can be analyzed independently. Tile-level predictions are subsequently aggregated to obtain slide-level inferences, enabling efficient learning while preserving local contextual detail that is critical for distinguishing subtle histomorphologic patterns ~\cite{gadermayr_multiple_2024}. This tiling paradigm underpins a range of downstream methods, including weakly supervised models \cite{krishnan_self-supervised_2022} and multiple-instance learning (MIL)\cite{lu_data-efficient_2021}, and has become a practical standard for translating WSIs into reproducible, quantitative biomarkers. The most recent paradigm shift in computational pathology approaches has been the development of foundation models (FMs), which derive information-rich representations via self-supervised learning on large-scale collections of WSIs\cite{waqas_revolutionizing_2023}. FMs have been developed in both tile- and slide-level variants, where slide-level FMs act as MIL aggregators of tile-level FM features. This provides multi-scale solutions for detailed computational analysis of histopathology images.

Tile-level FMs have been shown to yield robust, scale-invariant localized representations which can be optimized to yield enhanced performance in tasks including survival prediction, image retrieval, and tissue classification \cite{lu_visual_language_2024,xiang_visionlanguage_2025,vorontsov_foundation_2024,xu_whole-slide_2024,bioptimus_hoptimus1_2025}.
In parallel, slide-level FMs \cite{ding_multimodal_2024,wang_pathology_2024,noauthor_mahmoodlabmadeleine_2025}, offer hierarchical encoding and context-rich global embeddings based on aggregating information from across entire slides. % derived from tile embeddings and slide-level feature aggregators. 
Similarities in the training paradigms, pre-training data, or architectures of FMs suggest they may capture overlapping information related to tissue architecture or appearance. 
All pathology FMs generate representations using similar pre-training objectives\cite{krishnan_self-supervised_2022, huang_self-supervised_2023}, though specific models such as CONCH \cite{lu_visual_language_2024} and MUSK \cite{xiang_visionlanguage_2025} integrate multi-modal tasks into pre-training, potentially adding unique information and context to model embeddings. %Similarities in model embeddings could also originate from similarities in  institutional characteristics (e.g., staining and procedural differences) \cite{lin_unveiling_2025, bommasani_opportunities_2022,schneider_foundation_2024}. 
Additionally, many FMs use similar pre-training data from large public repositories \cite{fedorov_nci_2021}. 
%Table \ref{tab:encoders})\cite{han_survey_2023} summarizes key aspects of several popular pathology FMs, indicating the wide usage of vision transformers (ViT) in this context. 

However, recent studies have also reported that downstream models trained on different FM representations demonstrate complementary performance in their prediction scores as well as attending to different regions on a slide level~\cite{neidlinger_benchmarking_2025}. Neidlinger et al benchmarked the performance of different FMs and found that specific combinations of of FM predictions outperformed individual FMs. Zhao et al. demonstrated that an ensemble of classifiers trained using different FMs outperforms models trained on individual FMs~\cite{zhao_uncertainty-aware_2025}. This prompts two critical questions: (1) to what degree are the underlying FM embeddings unique and how much information is shared between diversely trained models (2) can we leverage multiple FMs to construct a unified embedding space that retains complementary information from different FMs but prunes out feature redundancy. To our knowledge, there has not yet been a detailed study of what specific information FM embeddings encapsulate relative to each other, or how to exploit their complementarity.  %how can the complementary information from each FM be optimally combined into a unified embedding space? 

A few obstacles exist in designing a unified embedding space between multiple FMs. The observed similarity in embedding representations across FMs inherently constrains the utility of directly concatenating their embeddings, since representational redundancy significantly complicates downstream analytical tasks \cite{dalvi_analyzing_2020, zollikofer_beyond_2024,tsukagoshi_redundancy_2025, you_fret_2025, el-manzalawy_min-redundancy_2018}. MIL frameworks, the  most common neural network subtype used in slide-level pathology prediction, are particularly susceptible to learning spurious correlations since identifying the most important tiles across a slide in an MIL framework requires parsing thousands of features across tens of thousands of tiles, but where only a small subset of tiles contain information necessary to predict the slide level label \cite{he_clustering-based_2020}. Any redundancy in feature representations would thus further compound the difficulty in identifying these few vital tiles in prediction tasks . 
Among the different MIL frameworks, the Clustering-constrained Attention Multiple-instance learning (CLAM) \cite{lu_data-efficient_2021} model is particularly prevalent, largely due to its ability to model instance-level variability through a gated attention mechanism. Nevertheless, recent studies indicate that CLAM's predictive accuracy may decline notably when confronted with highly correlated or redundant feature embeddings \cite{gadermayr_multiple_2024}. This underscores the necessity for a thorough, quantitative assessment of redundancy among leading pathology FMs and the development of sophisticated embedding fusion strategies to leverage the unique information inherent to each model for clinical predictions. 

Cancer grading and staging represents one such critical histopathological endpoint in oncologic diagnosis and prognostication\cite{goldenberg_new_2019}, which pose unique technical challenges\cite{klager_application_2025}. The primary challenge is that grade and stage assignments are made at the whole‐slide level rather than the patch of region level, reflecting aggregate morphological patterns of nuclear atypia, architectural disarray, and mitotic activity \cite{samaratunga_isup_2014}. Consequently, any computational framework must account for the vast scale of WSIs containing granular, cell‐level heterogeneity, either by directly generating slide‐level embeddings or by fusing tile‐level representations via MIL paradigms.

Early attempts to leverage slide‐based FM embeddings for grading have shown promise \cite{ding_multimodal_2024, wang_pathology_2024, noauthor_mahmoodlabmadeleine_2025}: by mapping an entire WSI into a single latent vector, models can capture global tissue context, stromal–tumor interactions, and field‐effect changes. Yet these slide‐level vectors could obscure critical focal features—microvascular proliferation, high‐grade clusters, or localized necrosis—that drive grade distinctions, potentially reducing efficacy for certain tasks. In parallel, tile‐level MIL schemes offer more granular sensitivity in capturing local grade‐defining cues. However, naïve MIL formulations risk diluting rare but diagnostically decisive tiles among hundreds of less informative ones. Tasks like cancer grading are particularly vulnerable to this, as small regions of high grade cancer can be decisive factors for slide level labels. Inappropriate attention or false negatives at the tile level can thus lead to false negatives at the slide-level in high‐grade cases\cite{gadermayr_multiple_2024}. This suggests the need to identify important information across an intelligent combination of diversely trained FMs, either at the tile- or slide-level, to not dilute diagnostically relevant signatures for optimal clinical outcome prediction.

\subsection{Study Goal}
In this study, we present a novel information-driven, intelligent fusion approach for foundation model embeddings, as well as a systematic evaluation of FM efficacy for cancer grading and staging. %By removing correlated features, we retain a discriminative signal while suppressing noisy redundant information from multiple FMs. \
Our approach is validated via digital pathology from across three disease contexts (kidney, prostate, and rectal cancer) as well as for both tile level and slide level models. We seek to explain the performance of such models by comprehensively analyzing and quantifying embedding redundancy among eight prominent pathology foundation models, observing and quantifying FM attention at the tile-level, as well as comparing the attention maps of downstream MIL models to specific tissue regions (healthy tissue, tumor) to interrogate the biological basis of foundation model-driven classifiers. %This study provides the first comprehensive view of FM fusion techniques, and provides a novel approach to optimally fuse diversely trained FM signatures. 

% Insert table of foundation models with all of their pertinent details
\section{Methods}

   \begin{figure} [ht]
   \begin{center}
   \begin{tabular}{c} %% tabular useful for creating an array of images 
   \includegraphics[width=\linewidth]{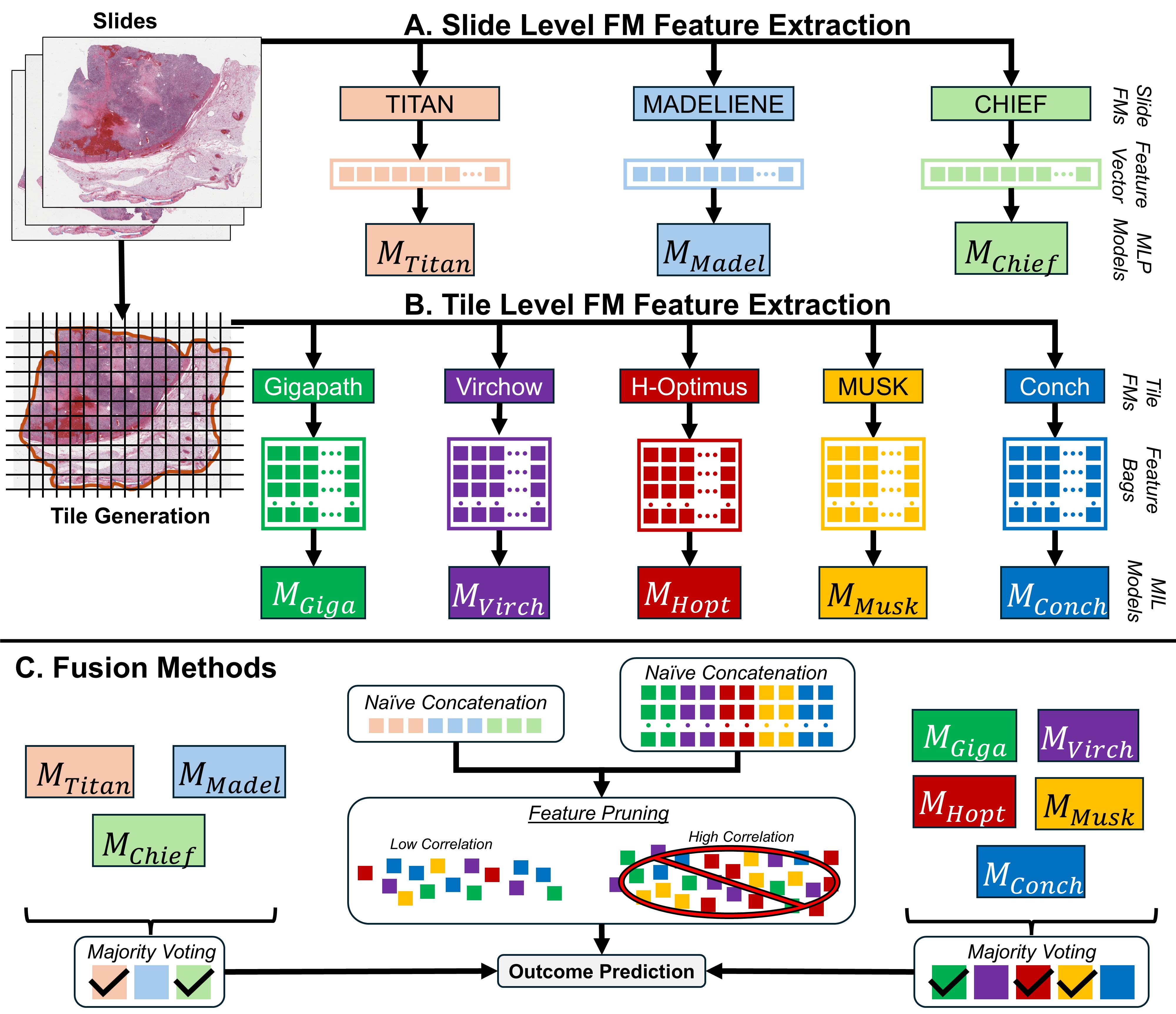}
   \end{tabular}
   \end{center}
   \caption{ 
   { \label{fig:workflow} Workflow for the study. (A) Slide-level foundation models were used to extract a single feature vector per slide. Multi-layered perceptons were trained to predict pathological characteristics using these features. (B) Tile level foundation models were used to extract a vector from each tile in the slide, resulting in a feature bag per slide. These feature bags were used to train multiple instance learning models to predict pathological characteristics. (C) Three fusion methods were implemented for both slide and tile level foundation models: (1) Majority voting (2) Naive feature concatenation (3) Intelligent fusion. %All three methods are used to develop predictions for the same outcomes and compared against each other and individual foundation models.
   }}
   \end{figure} 
   
\subsection{Data Description}

Diagnostic  hematoxylin and eosin (H\&E)-stained WSIs were curated from three publicly available pathology datasets from The Cancer Genome Atlas (TCGA).

\begin{itemize}
    \item Kidney Cancer: 519 hematoxylin and eosin (H\&E)-stained WSIs from 242 patients from TCGA Kidney Renal Clear Cell Carcinoma (TCGA-KIRC) collection. Available kidney cancer grades were based on ISUP guidelines, which were merged such that Grades 1–2 corresponded a “low-grade” category and Grades 3–4 into a “high-grade” category. 
    \item Prostate Cancer: 490 diagnostic H\&E-stained WSIs from 490 patients from TCGA Prostate Adenocarcinoma (TCGA-PRAD) collection. Available prostate cancer Gleason grades were subgrouped such that Gleason scores 1-6 corresponded to "low-grade" and 7-10 into "high-grade". 
    \item Rectal Cancer: 200 diagnostic H\&E-stained WSIs from 200 patients from TCGA Rectum Adenocarcinoma (TCGA-READ) collection. Rectal cancer pathology staging based on the AJCC guidelines were then grouped such that stages I-II were "low-stage" and III-IV were "high-stage".
\end{itemize}
 
A patient-stratified three-fold cross-validation scheme was implemented for each dataset, where %
%,  ($D^d_{te}$, where $d$ is the target disease). 
each fold included a training set ($D_{tr}$, 90\%) and  an internal validation set ($D_{iv}$, 10$\%$).
Performance of optimized models after cross validation was determined on a separate hold-out test set, comprising 10$\%$ of the total patients per cohort.

\subsection{Feature Extraction}
Tile embeddings were extracted using the TRIDENT framework \cite{Zhang2025Trident} (Fig \ref{fig:workflow}A-B). TRIDENT automatically identifies usable tissue using HEST, a pre-trained U-net segmenter optimized to identify tissue on H\&E slides\cite{jaume2024hest}. Tiles were generated from  HEST identified tissue regions with no overlap. Features were extracted using five tile level FMs: Conch v1.5\cite{lu_visual_language_2024}, MUSK\cite{xiang_visionlanguage_2025}, Virchow2\cite{vorontsov_foundation_2024}, H-optimus1\cite{bioptimus_hoptimus1_2025}, and Prov-gigapath\cite{xu_whole-slide_2024}; at the magnification and patch sizes recommended for each FM (Tab. \ref{tab:encoders}) as well as a 512x512 px normalized tile size at 20x magnification for tile level feature concatenation. Slide-level models were applied only at their recommended tile size and magnification (TITAN \cite{ding_multimodal_2024}, CHIEF \cite{wang_pathology_2024}, MADELEINE \cite{noauthor_mahmoodlabmadeleine_2025}). FM architectures, training details, and characteristics are summarized in Table \ref{tab:encoders}. All analyses were executed using a single NVIDIA L40 S GPU with 48 GB VRAM and 32 GB system RAM.

\begin{table*}[t]
\centering
\begingroup
\scriptsize
\setlength{\tabcolsep}{2.5pt}
\renewcommand{\arraystretch}{1.1}

% --- Tunable knob: shrink factor ( >1.00 = slightly wider internal layout ⇒ slight downscale ) ---
\newcommand*\tableshrinkfactor{1.10} % try 1.05–1.15 depending on how much you want to shrink

% Force a slight downscale while preserving the original wrapping decisions:
\adjustbox{width=\textwidth,center}{%
  \begin{minipage}{\tableshrinkfactor\textwidth}
    \begin{tabularx}{\linewidth}{l *{8}{Y}}
      \toprule
      \textbf{Property} & \textbf{Conch v1.5} & \textbf{MUSK} & \textbf{Virchow2} & \textbf{H-Optimus1} & \textbf{Prov-Gigapath} & \textbf{CHIEF} & \textbf{Madaleine} & \textbf{TITAN} \\
      \midrule
      Image Target        & Patch & Patch & Patch & Patch & Patch & Slide & Slide & Slide \\
      Architecture        & ViT-L/16 & BEiT-3 & ViT-H/14 & ViT-Custom & ViT-G/14 & AttMIL & AttMIL & ViT-B/CONCH (6L) \\
      Training Images     & 1.17 million patches & 50 million patches & 3.1 million patches & 1 million slides & 1.3 billion patches & 60.5 thousand slides & 16.8 thousand slides & 335.6 thousand slides \\
      Pre-training Method & CoCa & MIM/BLIP & DINOv2 & Unknown & DINOv2 & Anatomical Site Alignment & Multi-stain Alignment & CoCa \\
      Trained Patch Size  & 512$\times$512 & 384$\times$384 & 224$\times$224 & 256$\times$256 & 256$\times$256 & N/A & N/A & N/A \\
      Trained Magnification & 20$\times$ & 20$\times$ & 20$\times$ & 20$\times$ & 20$\times$ & 10$\times$ & 10$\times$ & 20$\times$ \\
      Embedding Dimension & 512 & 1024 & 2560 & 1536 & 1536 & 768 & 512 & 768 \\
      \bottomrule
    \end{tabularx}
  \end{minipage}%
}%
\endgroup
\caption{\label{tab:encoders}Summary of the pathology foundation models explored in this study. }
\end{table*}

\subsection{Experiment 1: Examination of fusion strategies for foundation models toward tumor characterization}
Independent models were trained for all three diseases: kidney cancer, prostate cancer, and rectal cancer. %where $n$ denotes the source of model features and $d$ is the target disease. 
A gated-attention CLAM \cite{lu_data-efficient_2021} model was trained separately for each tile level FM:  Conch v1.5\cite{lu_visual_language_2024} ($M_{Conch}$), MUSK\cite{xiang_visionlanguage_2025} ($M_{Musk}$), Virchow2\cite{vorontsov_foundation_2024} ($M_{Virch}$), H-optimus1\cite{bioptimus_hoptimus1_2025} ($M_{Hopt}$), and Prov-gigapath\cite{xu_whole-slide_2024} ($M_{Giga}$). Training used attention layers with 256 dimensions, dropout rate of 0.5, and Adam optimizer with learning rate $2\times10^{-4}$ and weight decay of $1\times10^{-5}$. Training employed one WSI per batch and was halted upon observing no improvement in validation loss after 20 epochs, with a maximum cap of 200 epochs. For slide level FMs, a six-layer multi-layer perceptrons (MLPs) were trained with the same settings as above for TITAN \cite{ding_multimodal_2024} ($M_{Titan}$), CHIEF \cite{wang_pathology_2024} ($M_{Chief}$), and MADELEINE \cite{noauthor_mahmoodlabmadeleine_2025} ($M_{Madel}$).

%\subsubsection{Fusion of Foundation Model Embeddings}

FM embeddings and decisions were integrated in the following ways, with the superscript $\psi$ denoting slide-level integration and $\tau$ corresponding to tile-level integration: 
\begin{itemize}
    \item Majority voting: Slide-level predictions from the five tile-level CLAM models were combined via majority vote, denoted $M^{\tau}_{Maj}$. Slide-level predictions were similarly combined from the three slide-level MLP models via majority vote, denoted $M^{\psi}_{Maj}$.
    %based on selecting the model with the highest F1 score on the internal validation set across all cross-validation folds.
    \item Naive feature fusion: FM embeddings were concatenated into a single, unified space, denoted $M^{\tau}_{Con}$ for tile level models and $M^{\psi}_{Con}$ for slide level models. 
    \item Intelligent fusion: $M^{\tau}_{Con}$ and $M^{\psi}_{Con}$ underwent a feature reduction process by first ranking FM features based on significant differences between classes (e.g. low vs high grade) and then pruning FM features which exhibited a Pearson correlation of greater than $\theta$ with any higher-ranked features. Pruned feature sets were compiled by varying $\theta = {0.1, 0.2, 0.3, 0.4, 0.5, 0.7}$, with models trained separately for the resulting feature set at each threshold. This yielded compact, minimally redundant FM feature sets, which were each used to train separate CLAM models, denoted $M^{\tau}_{IF}$ and $M^{\psi}_{IF}$), respectively.
\end{itemize}
%Individual FM models as well the fusion FM approaches were evaluated and trained via a single iteration of patient-stratified three-fold cross-validation with the same data splits as single FM based CLAM models used above. %Performance was quantified in terms of classification AUC, sensitivity, specificity and F1 score within the internal validation set as well as the external test cohort.

%All models were evaluated using the methods described in Section \ref{sec:stats}.

Performance was quantified in terms of classification AUC, sensitivity, specificity and F1 score within the internal validation set as well as the external test cohort for each model and each disease classification task. Statistical comparisons between models were conducted with bootstrapping using 50 iterations of 80\% of the holdout testing split. $p$-values were corrected for multiple comparisons. 

\subsection{Experiment 2: Evaluating the similarity of foundation model embeddings}
To characterize the similarity of the embedding spaces defined by each FM, 50,000 tiles were randomly sampled from each disease cohort followed by computing the following metrics between each pair of FM embeddings: 

\begin{itemize}
    \item Centered Kernel Alignment (CKA) \cite{kornblith_similarity_2019}: Measures the similarity of two embedding sets by comparing their centered Gram (kernel) matrices via the Hilbert–Schmidt Independence Criterion. CKA is invariant to orthogonal transformations and isotropic scaling, and reliably matches corresponding layers across different network initializations.
    \item Singular-vector canonical correlation analysis (SVCCA) \cite{raghu_svcca_2017}: Uses singular value decomposition (SVD) to project each embedding set onto its top singular‐vector subspace, then applies Canonical Correlation Analysis to those subspaces, reporting the average canonical correlation—thus quantifying alignment of principal directions in two representations.
    \item Orthogonal Procrustes Distance (OPD\cite{andreella_procrustes-based_2023}: Reduces both embedding sets (e.g. via PCA) to the same dimension and finds the optimal orthogonal rotation that minimizes their Frobenius‐norm difference; the resulting normalized residual (Procrustes distance) captures “shape” dissimilarity between point clouds
    \item Jaccard Index of k-Nearest Neighbor (k-NN) Overlap \cite{tavares_measuring_2024} : For each sample, finds its k nearest neighbors in each embedding space and computes the Jaccard similarity (intersection over union) of those neighbor sets. Averaging these per-sample scores yields a fine-grained, neighborhood-based measure of local embedding agreement.
    \item Cross Prediction Ridge Regression (RR) \cite{kornblith_similarity_2019}: Fits two ridge‐regularized linear maps (Y to X and X to Y) with L2 penalty and reports the proportion of variance in one embedding explained by the other. The resulting $R^2$ scores quantify how well one feature set linearly predicts the other.
\end{itemize}

CKA, SVCCA, OPD, and RR are targeted to measuring global similarity of embedding spaces, whereas k-NN measures local similarity of specific embeddings. Global similarity is a measure of the alignment of the embeddings representations across the entire dataset, while local similarity is the alignment of specific small neighborhoods within the entire network of samples. These metrics also span spectral methods (SVCCA, OPD) and graph based methods (k-NN), providing a comprehensive view of model embedding similarity. For all metrics except OPD, higher values indicate greater similarity between the compared embeddings. For OPD, higher values indicate greater distance between two embedding spaces, and therefore less similarity. These metrics were computed and compared across all three disease spaces for both tile and slide level FMs, separately. 

\subsection{Experiment 3: Evaluating foundation model attention maps and unsupervised clusterings}
To identify the specific tissues that drove foundation model performance, model attention of different tile-level foundation models and their downstream slide-level CLAM models were compared. 

\subsubsection{Slide-level attention}
%Regions on pathology slides that were highly attended by the various CLAM models trained in this study. 
Attention of the different slide-level CLAM models were compared to examine how different FM features drive attention towards different regions on pathology slides. To accomplish this, tile attention values were extracted from each kidney CLAM model and mapped onto the original pathologic slides in order to examine attention across regions encompassing multiple tiles. Attention of $M_{Conch}$, $M_{Virch}$, $M_{Musk}$, $M_{Giga}$, $M_{Hopt}$, $M^\tau_{Con}$, and $M^\tau_{IF}$ models were compared by calculating the percentage of tumor and normal tissue that received attention from each model and for each dataset separately. A tile was considered attended to if its attention value surpassed a given percentile threshold. Attention thresholds of 25th, 50th, 60th, 70th, 80th, and 90th percentiles were tested where tiles at higher percentile attention  indicate those regions were weighted highly for grade prediction. 

\subsubsection{Tile-level attention}
In contrast to slide-level attention, tile-level attention maps indicate the specific pixel regions that provide information used to calculate tile-level FM embeddings. To examine the differences in tile-level attention between different foundation models, attention maps were extracted from the first attention layer of each ViT architecture when those models were applied to kidney cancer WSIs. Since the FM MUSK uses a BEiT3 architecture instead of ViT,   attention values were extracted directly from the multi-head attention module. The overlap of high attention regions was calculated by creating binary masks at the 50th, 70th, and 90th percentile of attention between each pair of FMs using Dice score. Dice score is a measure of overlap between two regions where 0 indicates no overlap and 1 indicates perfect overlap. Overlap at 50th percentile would indicate similar regions receive some attention to derive features, while overlap at 90th percentile indicates features are derived from similar specific biological targets (cells, tissue primitives). 

\subsubsection{Measuring tissue region clustering ability of foundation model embeddings}
To measure the ability of an FM representation to cluster distinct tissue regions (tumor, benign tissue) in an unsupervised manner, tile embeddings were accumulated across the dataset and projected into 2D-space using t-Distributed Stochastic Neighbor Embedding (t-SNE). Benign and tumor tissues were considered as two distinct clusters in this space, where well defined clusters indicate FM signatures capture distinct appearance differences between tissue types. Clustering accuracy was measured by silhouette coefficient, which measures both the ability of a feature set to separate tissue types as well as the compactness of each tissue cluster. Silhouette coefficient ranges from -1 (poor clustering) to +1 (perfect clustering). Tumor and benign compactness were also measured separately by calculating the median distance between a class instance (tile embedding) and its cluster centroid. This process was completed for the intelligently fused signature, naive fusion methods, as well as individual FMs. Tissue cluster compactness ranges from 0 (very dispersed) to +1 (very compact). Clustering metrics were compared between feature sets using Wilcoxon ranksum tests after fifty bootstrap iterations with 80\% of the tiles across the dataset. 
% \subsection{Experiment 4: Biological Underpinning of Foundation Model Embeddings}
%...something we will probably save for a future paper

% \subsection{Statistical Analysis}
% \label{sec:stats}

\section{Results}

\subsection{Experiment 1: Foundation Model Embedding Fusion for Characterizing Cancer}

% Tile level model performance_____________________________________________
\begin{table*}[!t]
\captionsetup{font=small}
\caption{\label{tab:tile_performance}Performance of MIL-CLAM models trained with different tile-level foundation-model embeddings as well as fusion methods (majority vote, concatenation, intelligent fusion). Best performing model within each disease block and column are bolded; ties are bolded. \(M_{\ast}\) denotes the model instantiated with the corresponding embedding (e.g., \(M_{\text{Conch}}\)).}
\centering
\setlength{\tabcolsep}{5.5pt}
\renewcommand{\arraystretch}{1.25}
\resizebox{\textwidth}{!}{%
\begin{tabular}{@{}l l l cccc cccc@{}}
\toprule
 & & & \multicolumn{4}{c}{Internal Validation} & \multicolumn{4}{c}{External Testing} \\
\cmidrule(lr){4-7}\cmidrule(l){8-11}
Disease & Embedding & Model & AUC & Sen & Spe & F1 & AUC & Sen & Spe & F1 \\
\midrule[1.1pt]
\multirow{8}{*}{Kidney}
 & Conch          & $M_{Conch}$ & \textbf{0.91} & \textbf{0.86} & \textbf{0.70} & \textbf{0.78} & 0.75 & 0.53 & \textbf{0.90} & 0.68 \\
 & MUSK           & $M_{Musk}$  & 0.81 & 0.79 & 0.60 & 0.70 & 0.81 & 0.80 & 0.60 & 0.70 \\
 & Virchow        & $M_{Virch}$ & 0.82 & 0.79 & 0.50 & 0.64 & \textbf{0.85} & 0.80 & 0.60 & 0.70 \\
 & H-Optimus1     & $M_{Hopt}$  & 0.84 & \textbf{0.86} & 0.60 & 0.73 & 0.80 & \textbf{0.93} & 0.50 & 0.72 \\
 & Prov-Gigapath  & $M_{Giga}$  & 0.86 & 0.79 & 0.50 & 0.64 & 0.82 & 0.80 & 0.80 & 0.79 \\
 & Fusion (Majority Vote)             & $M^\tau_{Maj}$   & --   & --   & --   & --   & \textbf{0.85} & 0.67 & 0.63 & 0.69 \\
 & Fusion (Concatenation)             & $M^\tau_{Con}$ & 0.69 & 0.64 & 0.60 & 0.62 & 0.69 & 0.70 & 0.60 & 0.63 \\
 & Fusion (Pruning)            & $M^\tau_{IF}$  & 0.79 & 0.79 & 0.50 & 0.64 & 0.84 & 0.80 & \textbf{0.90} & \textbf{0.84} \\
\midrule[1.1pt]
\multirow{8}{*}{Prostate}
 & Conch          & $M_{Conch}$ & 0.94 & 0.83 & 0.88 & 0.86 & 0.94 & 0.71 & \textbf{0.97} & 0.84 \\
 & MUSK           & $M_{Musk}$  & \textbf{0.97} & 0.93 & 0.92 & \textbf{0.92} & 0.96 & 0.80 & 0.96 & 0.88 \\
 & Virchow        & $M_{Virch}$ & 0.94 & 0.81 & 0.88 & 0.85 & 0.93 & 0.85 & \textbf{0.97} & 0.91 \\
 & H-Optimus1     & $M_{Hopt}$  & 0.89 & 0.69 & 0.81 & 0.75 & 0.93 & 0.79 & \textbf{0.97} & 0.86 \\
 & Prov-Gigapath  & $M_{Giga}$  & 0.91 & 0.83 & \textbf{0.96} & 0.90 & 0.94 & 0.88 & 0.87 & 0.85 \\
 & Fusion (Majority Vote)             & $M^\tau_{Maj}$   & --   & --   & --   & --   & 0.97 & 0.88 & \textbf{0.97} & 0.92 \\
 & Fusion (Concatenation)            & $M^\tau_{Con}$ & 0.92 & \textbf{0.94} & 0.85 & 0.88 & \textbf{0.99} & 0.88 & \textbf{0.97} & 0.92 \\
 & Fusion (Pruning)            & $M^\tau_{IF}$  & 0.93 & 0.88 & 0.85 & 0.85 & 0.98 & \textbf{0.92} & \textbf{0.97} & \textbf{0.94} \\
\midrule
\multirow{8}{*}{Rectal}
 & Conch          & $M_{Conch}$ & 0.86 & 0.85 & 0.75 & 0.80 & 0.90 & 0.90 & 0.50 & 0.71 \\
 & MUSK           & $M_{Musk}$  & 0.86 & 0.78 & 0.86 & 0.78 & 0.94 & \textbf{0.95} & 0.67 & 0.85 \\
 & Virchow        & $M_{Virch}$ & 0.94 & 0.90 & 0.83 & 0.87 & 0.94 & \textbf{0.95} & \textbf{0.83} & \textbf{0.89} \\
 & H-Optimus1     & $M_{Hopt}$  & 0.99 & \textbf{0.95} & 0.71 & 0.85 & \textbf{0.97} & 0.90 & \textbf{0.83} & 0.85 \\
 & Prov-Gigapath  & $M_{Giga}$  & 0.96 & \textbf{0.95} & 0.75 & 0.86 & 0.91 & 0.86 & 0.67 & 0.75 \\
 & Fusion (Majority Vote)             & $M^\tau_{Maj}$   & --   & --   & --   & --   & \textbf{0.97} & 0.94 & 0.67 & 0.86 \\
 & Fusion (Concatenation)            & $M^\tau_{Con}$ & \textbf{1.00} & \textbf{0.95} & \textbf{0.83} & \textbf{0.93} & 0.95 & \textbf{0.95} & 0.67 & 0.82 \\
 & Fusion (Pruning)            & $M^\tau_{IF}$  & 0.98 & 0.90 & \textbf{0.92} & 0.90 & \textbf{0.97} & \textbf{0.95} & \textbf{0.83} & \textbf{0.89} \\
\bottomrule
\end{tabular}%
}
\end{table*}

\begin{figure}[t]
  \centering
  \computeimgmaxheight
  \includegraphics[width=\textwidth,height=21cm,keepaspectratio]{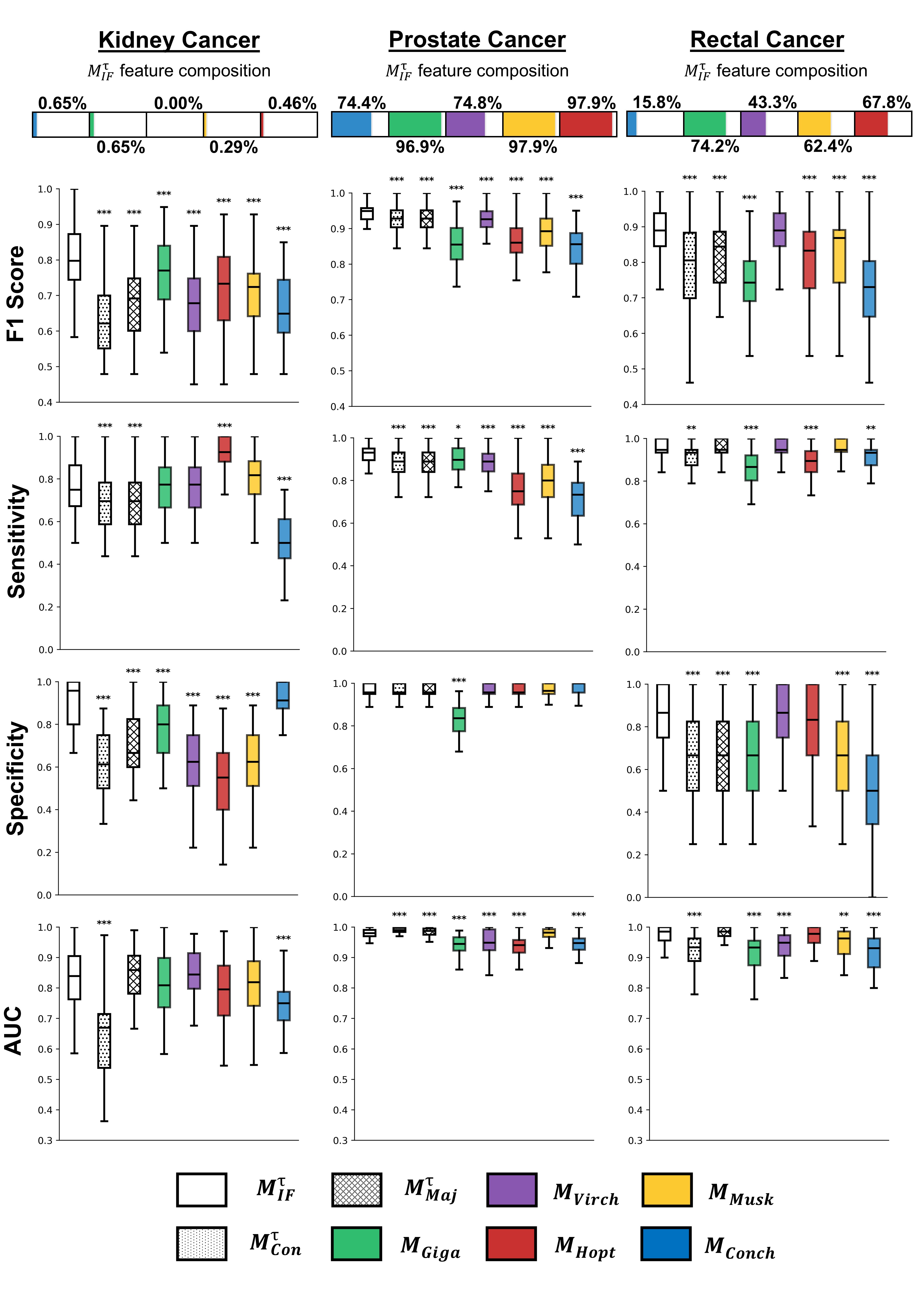}
  \caption{\label{fig:tile_performance} Performance and statistical comparison of MIL-CLAM models trained using tile level foundation model features. Significance as determined by wilcoxen log-rank tests between the intelligent fusion method and other models is indicated above each respective box ($* (p=0.5/N)$, $** (p=0.01/N)$, $*** (p=0.01/N)$, where $N$ is the number of comparisons. Thermometer plot shows percentage of features from each FM included in feature set used to train $M_{IF}^\tau$.}
\end{figure}

%Figure \ref{fig:tile_performance} summarizes the performance of MIL-CLAM models trained with tile-level FM embeddings. Intelligent fusion of tile-level FM features yielded consistent, statistically significant improvements in hold-out performance across all three disease cohorts compared with models trained on individual FM embeddings and with naïve fusion schemes. In kidney and prostate cancer, intelligent fusion models ($M_{IF}^\tau$) achieved the highest F1 scores on holdout testing cohorts while in rectal cancer $M_{IF,T}^R$ outperformed all comparators except $M_{Virch}^R$. Across diseases, intelligent fusion models also exceeded majority-vote ensembles ($M_{Maj,T}^d$) and simple concatenation ($M_{Con,T}^d$), and demonstrated a favorable balance of sensitivity and specificity, typically matching or surpassing the best-performing single-FM models. AUC values for the intelligent fusion models were comparable to or higher than those of individual FMs and alternative fusion strategies in kidney and rectal cancer, with prostate cancer representing the only setting in which a non-pruned fusion configuration achieved a marginally higher AUC. Taken together, these findings suggest that correlation-guided pruning of tile-level FM embeddings provides a more informative and robust representation than either single-model features or uninformed multi-model fusion.

Figure \ref{fig:tile_performance} summarizes the performance of MIL-CLAM models trained with tile-level FM embeddings. Intelligent fusion of tile-level FM features yielded consistent, statistically significant improvements in hold-out performance across all three disease cohorts compared with classifiers trained on individual FM embeddings and with naïve fusion schemes. In kidney and prostate cancer, intelligent fusion classifiers at the tile- and slide-level achieved the highest F1 scores on holdout testing cohorts (kidney: 0.84, prostate: 0.94) while in rectal cancer the intelligent fusion classifier outperformed all comparators except Virchow (F1 of 0.89 for both). Across diseases, intelligent fusion classifiers also exceeded majority-vote ensembles and simple concatenation, as well as demonstrating a favorable balance of sensitivity and specificity, typically matching or surpassing the best-performing single-FM classifiers. AUC values for the intelligent fusion classifiers were comparable to or higher than those of individual FMs and alternative fusion strategies in kidney and rectal cancer, with prostate cancer representing the only setting in which a non-pruned fusion configuration achieved a marginally higher AUC. Taken together, these findings suggest that correlation-guided pruning of tile-level FM embeddings provides a more informative and robust representation than either single-model features or uninformed multi-model fusion.

\begin{table*}[!t]
\captionsetup{font=small}
\caption{\label{tab:slide_performance} Performance of multi-layered perceptron models trained using different slide-level foundation model embeddings as well as fusion methods (majority vote, concatenation, intelligent fusion). Best performing model within each disease block and column are bolded; ties are bolded. \(M_{\ast}\) denotes the model instantiated with the corresponding embedding (e.g., \(M_{\text{Titan}}\)).}
\centering
\setlength{\tabcolsep}{5.5pt}
\renewcommand{\arraystretch}{1.25}
\resizebox{\textwidth}{!}{%
\begin{tabular}{@{}l l l cccc cccc@{}}
\toprule
 & & & \multicolumn{4}{c}{Internal Validation} & \multicolumn{4}{c}{External Testing} \\
\cmidrule(lr){4-7}\cmidrule(l){8-11}
Disease & Embedding & Model & AUC & Sen & Spe & F1 & AUC & Sen & Spe & F1 \\
\midrule[1.1pt]
\multirow{6}{*}{Kidney}
 & TITAN      & $M_{Titan}$ & 0.69 & 0.79 & 0.50 & 0.73 & 0.67 & 0.65 & 0.65 & 0.68 \\
 & MADELEINE  & $M_{Madel}$ & 0.81 & 0.86 & 0.30 & 0.73 & 0.74 & 0.73 & \textbf{0.95} & \textbf{0.78} \\
 & CHIEF      & $M_{Chief}$ & 0.78 & 0.71 & \textbf{0.60} & 0.71 & 0.71 & 0.67 & 0.80 & 0.74 \\
 & Fusion (Majority Vote)        & $M^\psi_{Maj}$   & \textbf{0.89} & \textbf{0.93} & \textbf{0.60} & \textbf{0.84} & 0.75 & 0.73 & 0.65 & 0.68 \\
 & Fusion (Concatenation)        & $M^\psi_{Con}$ & 0.78 & 0.79 & 0.50 & 0.73 & 0.73 & 0.73 & 0.70 & 0.70 \\
 & Fusion (Pruning)        & $M^\psi_{IF}$  & 0.75 & 0.86 & 0.40 & 0.75 & \textbf{0.76} & \textbf{0.87} & 0.70 & \textbf{0.78} \\
\midrule[1.1pt]
\multirow{6}{*}{Prostate}
 & TITAN      & $M_{Titan}$ & 0.90 & \textbf{0.83} & \textbf{0.96} & \textbf{0.88} & 0.93 & 0.79 & 0.97 & 0.84 \\
 & MADELEINE  & $M_{Madel}$ & 0.89 & 0.75 & 0.88 & 0.77 & 0.94 & 0.88 & 0.92 & 0.88 \\
 & CHIEF      & $M_{Chief}$& 0.88 & 0.69 & 0.85 & 0.71 & 0.98 & 0.92 & 0.97 & 0.96 \\
 & Fusion (Majority Vote)         & $M^\psi_{Maj}$   & 0.93 & 0.81 & 0.88 & 0.81 & 0.97 & 0.83 & 1.00 & 0.93 \\
 & Fusion (Concatenation)        & $M^\psi_{Con}$ & \textbf{0.96} & 0.78 & \textbf{0.96} & 0.86 & 0.97 & 0.83 & \textbf{1.00} & 0.91 \\
 & Fusion (Pruning)        & $M^\psi_{IF}$  & 0.82 & 0.78 & 0.85 & 0.78 & \textbf{0.99} & \textbf{0.96} & 0.99 & \textbf{0.97} \\
\midrule[1.1pt]
\multirow{6}{*}{Rectal}
 & TITAN      & $M_{Titan}$ & 0.88 & \textbf{1.00} & 0.29 & 0.89 & 0.94 & \textbf{0.98} & 0.67 & 0.85 \\
 & MADELEINE  & $M_{Madel}$ & \textbf{0.98} & \textbf{1.00} & \textbf{0.75} & \textbf{0.93} & 0.93 & 0.91 & 0.67 & 0.79 \\
 & CHIEF      & $M_{Chief}$ & 0.72 & 0.95 & 0.29 & 0.86 & 0.97 & \textbf{0.98} & 0.67 & 0.83 \\
 & Fusion (Majority Vote)         & $M^\psi_{Maj}$   & 0.86 & 0.95 & 0.57 & 0.90 & \textbf{0.98} & \textbf{0.98} & \textbf{0.90} & \textbf{0.88} \\
 & Fusion (Concatenation)        & $M^\psi_{Con}$ & 0.88 & 0.94 & 0.29 & 0.85 & 0.95 & \textbf{0.98} & \textbf{0.90} & \textbf{0.88} \\
 & Fusion (Pruning)        & $M^\psi_{IF}$  & 0.83 & 0.89 & 0.43 & 0.84 & 0.96 & 0.95 & \textbf{0.90} & 0.86 \\
\bottomrule
\end{tabular}%
}
\end{table*}

\begin{figure}[t]
  \centering
  \computeimgmaxheight
  \includegraphics[width=\textwidth,height=21cm,keepaspectratio]{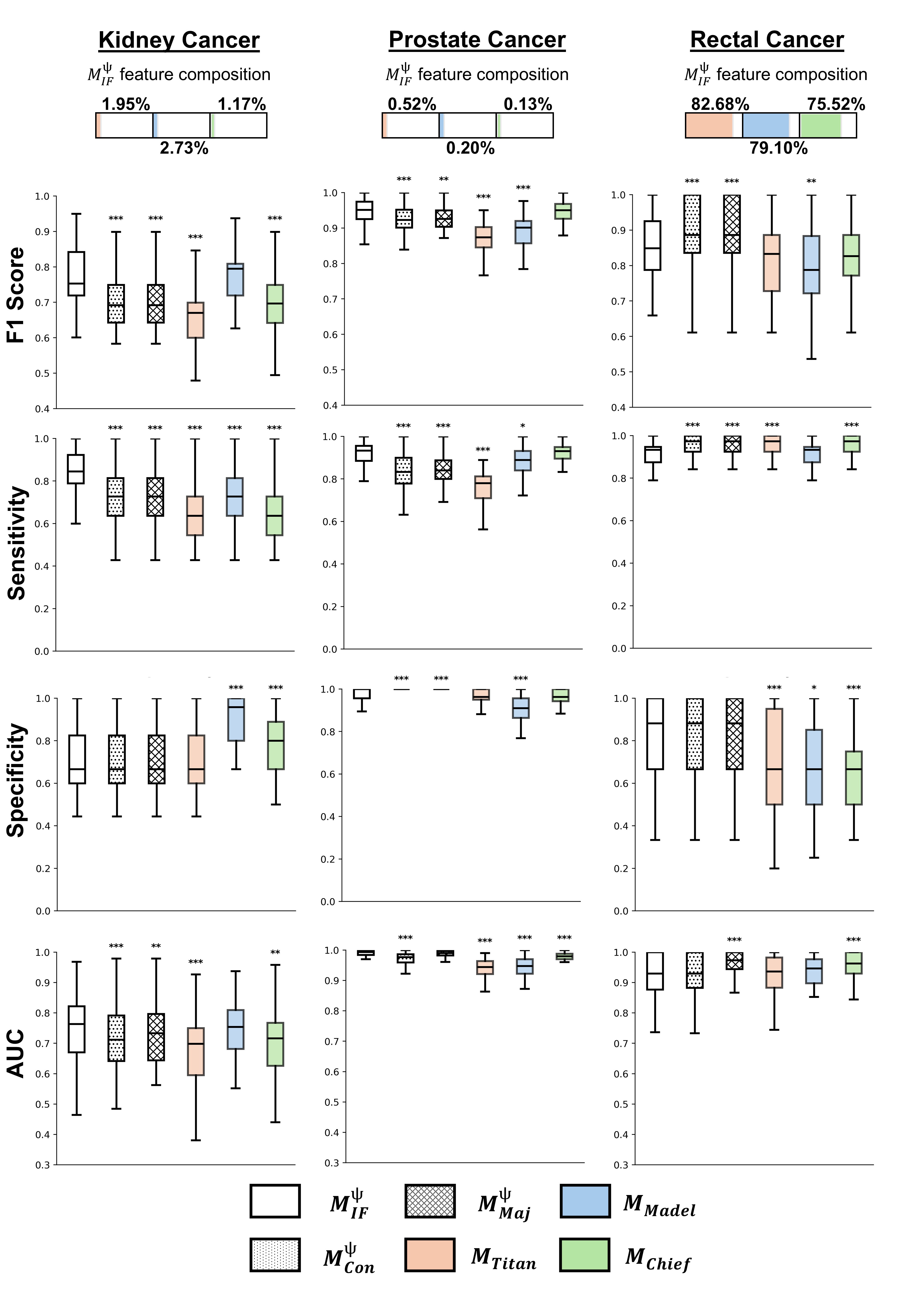}
  \caption{\label{fig:slide_performance} Performance and statistical comparison of multi-layer perceptrons (MLPs) trained using slide-level foundation models. Significance as determined by Wilcoxon log-rank tests between the intelligent fusion method and other models is indicated above each respective box ($* (p=0.5/N)$, $* (p=0.01/N)$, $* (p=0.01/N)$, where $N$ is the number of comparisons. Thermometer plot shows percentage of features from each FM included in feature set used to train $M_{IF}^\psi$.}
\end{figure}

%Figure \ref{fig:slide_performance} summarizes performance of MLPs trained with slide-level FM embeddings. Intelligent fusion of slide-level features yielded modest but consistent gains in kidney and prostate cancer. In kidney, the intelligent fusion model $M_{IF,S}^K$ achieved the highest F1 score (0.78) on holdout testing, comparable to the best individual FM model ($M_{Madel}^K$), with similar improvements reflected in AUC and sensitivity. In prostate cancer, $M_{IF,S}^P$ attained an F1 score of 0.97 on holdout testing, numerically outperforming all comparators but not significantly exceeding the strongest single-FM baseline, and exhibited high AUC and sensitivity with competitive specificity. By contrast, for rectal cancer stage prediction, simple fusion strategies outperformed both individual FM models and the intelligent fusion model across AUC, sensitivity, and F1 score, indicating that slide-level feature pruning is less beneficial in this setting than straightforward multi-model fusion.

Figure \ref{fig:slide_performance} summarizes performance of MLPs trained with slide-level FM embeddings. Intelligent fusion of slide-level features yielded modest but consistent gains in kidney and prostate cancer. In kidney cancer, the intelligent fusion classifier achieved the highest F1 score (0.78) on holdout testing, comparable to the best individual FM classifier MADELEINE (0.78), with similar improvements reflected in AUC and sensitivity. In prostate cancer, the intelligent fusion classifier attained an F1 score of 0.97 on holdout testing, numerically outperforming all comparators but not significantly exceeding the strongest single-FM baseline (CHIEF, 0.96), while exhibiting high AUC and sensitivity with competitive specificity. By contrast, for rectal cancer stage prediction, simple fusion strategies outperformed both individual FM classifiers and the intelligent fusion classifier across AUC, sensitivity, and F1 score, indicating that slide-level feature pruning is less beneficial in this setting.%than straightforward multi-FM fusion.

% Figure \ref{fig:slide_performance} shows performance and statistical comparisons for MLPs trained using slide level FM embeddings. Slide level FM feature pruning also facilitated some increase in performance on holdout testing cohorts ($D^K_{te}$, $D^P_{te}$, $D^R_{te}$) (Tab. \ref{tab:slide_performance}). $M_{Corr, S}^K$ exhibited a significantly improved F1 score of 0.78 compared to all other models and feature fusion methods while matching the performance of $M_{Madel}^K$ (Fig. \ref{fig:slide_performance}). Similarly, $M_{Corr, S}^P$ outperforms all models with an F1 score of 0.97, but is not significantly better than $M_{Chief}^K$ (0.96). Similar trends were observed for $M_{Corr, S}^K$ and $M_{Corr, S}^P$ for both AUC and Sensitivity. $M_{Corr, S}^P$ achieved competitive specificity (0.99), while $M_{Madel}^K$ (0.95) and $M_{Chief}^K$ (0.80) significantly outperformed $M_{Corr, S}^K$ (0.70). For rectal cancer stage prediction, simple fusion methods demonstrated significantly better performance compared to multiple individual models as well as $M_{Corr, S}^R$ for AUC, Sensitivity, and F1 score. 

% pruning across different correlation thresholds ____________________________

\begin{figure}
\begin{center}
\includegraphics[width=\linewidth]{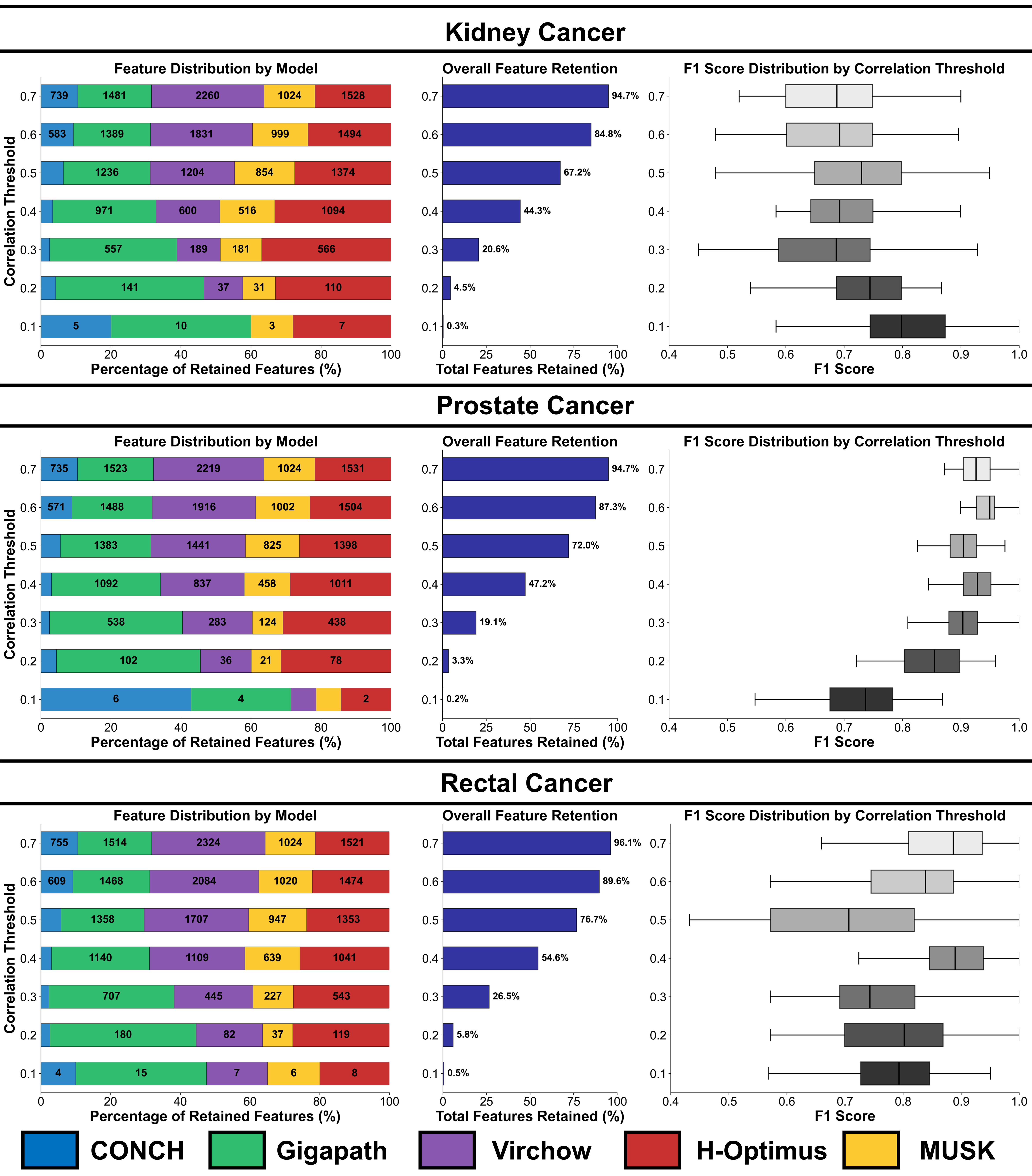}  % fig2 includes two images 
\end{center}
\caption 
{ \label{fig:flowchart}
\label{fig:tile_pruning} Feature selection, feature retention, and downstream model performance for intelligent fusion of tile-level foundation models. Numbers in the feature distribution bars correspond to the number of features chosen from that models at each specific correlation threshold. The size of each models bar represents the percentage contribution to the total feature vector. The overall feature retention represents the total percentage of features retained at each pruning threshold. Performance box plots summarize model F1 scores, with a darker shade of gray indicating a lower pruning threshold. } 
\end{figure} 

Figure \ref{fig:tile_pruning} depicts the evolution of feature retention and model performance as the correlation threshold for tile-level intelligent fusion is varied. Across all three diseases, relatively low thresholds remove the vast majority of features, with thresholds near 0.4 eliminating approximately half of all input FM features while setting the thresholds at 0.1 consistently prunes more than 99\% of FM features. Optimal performance on the hold-out cohorts was achieved at distinct thresholds for each disease (kidney: 0.1; prostate: 0.6; rectal: 0.4 by mean F1 score), suggesting disease-specific balances between pruning of redundant FM features and information content. The relative contribution of individual FMs to the retained feature set also followed characteristic patterns. Conch was preferentially preserved at very low thresholds in kidney and prostate cancer, contributing a substantially larger fraction of retained features than its proportion in naive concatenation, but its contribution reduced as the threshold increased. In contrast, features from Prov-Gigapath, H-Optimus, and Virchow generally became progressively underrepresented as the pruning threshold increased, with Virchow in some instances being almost entirely pruned out at the lowest thresholds. MUSK exhibited the opposite trend, contributing an increasing fraction of the retained features with higher thresholds and remaining fully preserved at the highest threshold tested across all three diseases. Taken together, these patterns indicate that correlation-based pruning preferentially retains a compact, disease- and FM-specific subset of embeddings. %while discarding large volumes of redundant information.

\begin{figure}
\begin{center}
\includegraphics[width=\linewidth]{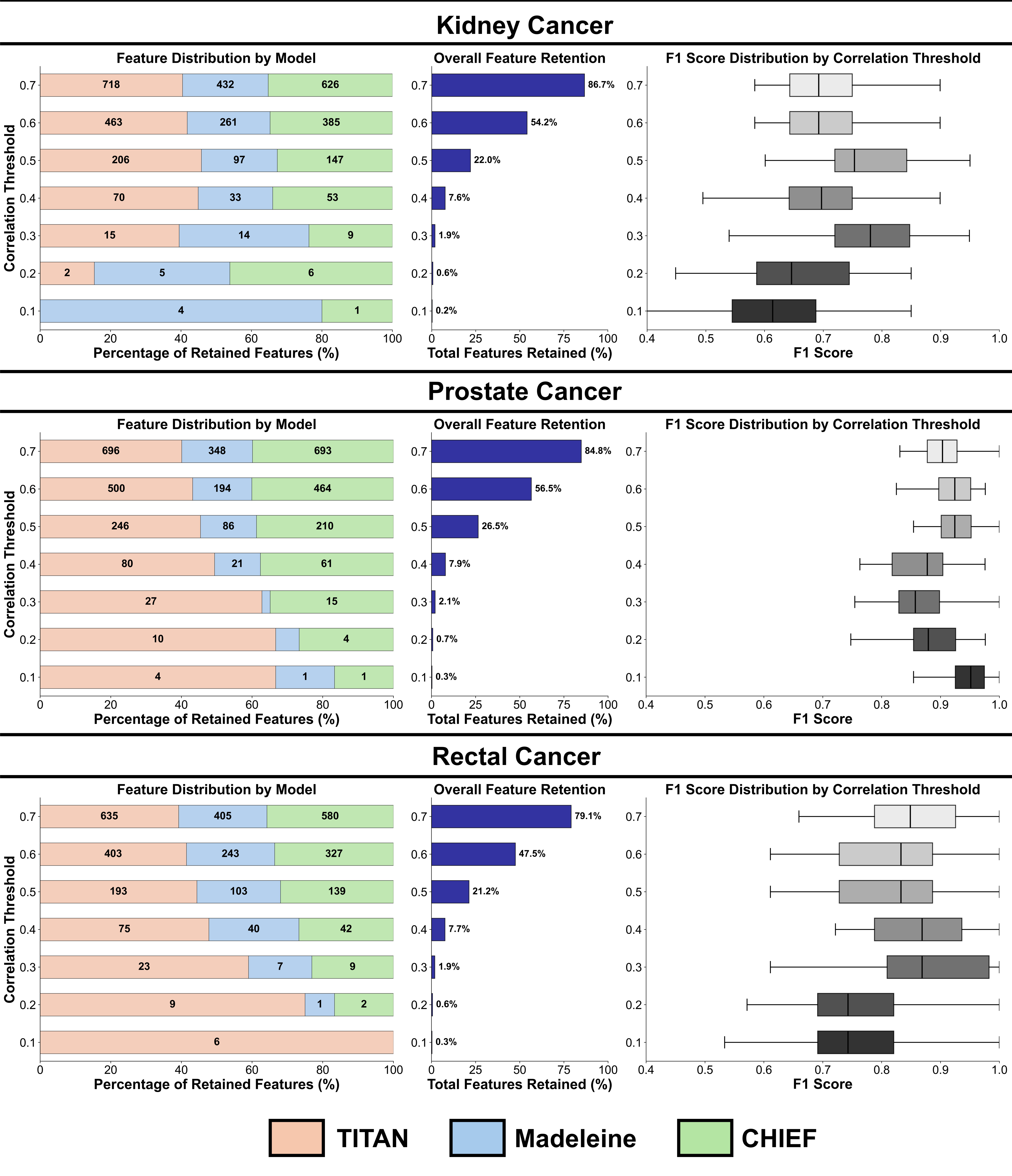}  % fig2 includes two images 
\end{center}
\caption 
{ \label{fig:flowchart}
\label{fig:slide_pruning} Feature selection, feature retention, and downstream model performance for intelligent fusion of slide-level foundation models. Numbers in the feature distribution bars correspond to the number of features chosen from that models each specific correlation threshold. The size of each models bar represents the percentage contribution to the total feature vector. The overall feature retention represents the total percentage of features retained at each pruning threshold. Performance box plots summarize model F1 scores, with a darker shade of gray indicating a lower pruning threshold. } 
\end{figure} 

Figure \ref{fig:slide_pruning} summarizes how correlation-based pruning of slide-level FM embeddings affects both feature retention and downstream performance in our intelligent fusion approach. Similar to tile-level results, optimal thresholds for outcome prediction differed by disease, with best mean F1 scores observed at thresholds of 0.3, 0.1, and 0.7 for kidney, prostate, and rectal cancer, respectively. Slide-level embeddings exhibited marked redundancy, such that even moderate thresholds removed a large proportion of features: fewer than 10\% of features were retained at a threshold of 0.4, and more than half were discarded by 0.6 across all three diseases. In contrast to the more stable patterns observed at the tile level, feature-selection behavior at the slide level was highly variable across diseases and FMs. For example, no TITAN features were selected at a threshold of 0.1 for kidney cancer, whereas at the same threshold only TITAN features were retained for rectal cancer. Consequently, the relative contribution of each FM to the pruned feature vector changed substantially between diseases, indicating that the redundancy of slide-level representations is strongly disease- and model-dependent.

% Figure \ref{fig:slide_pruning} shows the number of features chosen from each slide-level FM embedding at different correlation thresholds as well as the performance of the model trained at each threshold. The best performing thresholds on holdout testing for kidney, prostate, and rectal cancer outcome prediction were 0.3, 0.1, and 0.7 by mean F1 score, respectively. Slide level features were pruned at much higher rates compared to tile level features. Greater than 50\% of features are not retained in any disease until a threshold of 0.6, and less than 10\% of features are retained at a threshold of 0.4. This indicates substantial similarity in features across and within model embeddings. Feature selection patterns were also more variable for slide level compared to tile level. No TITAN features were selected at 0.1 for kidney cancer, but only TITAN features were selected at the same threshold for rectal cancer. Consequently, the relative percentage that different models encompassed of the pruned feature vector changes dramatically between different diseases. 

\begin{figure}
\begin{center}
\includegraphics[width=\linewidth]{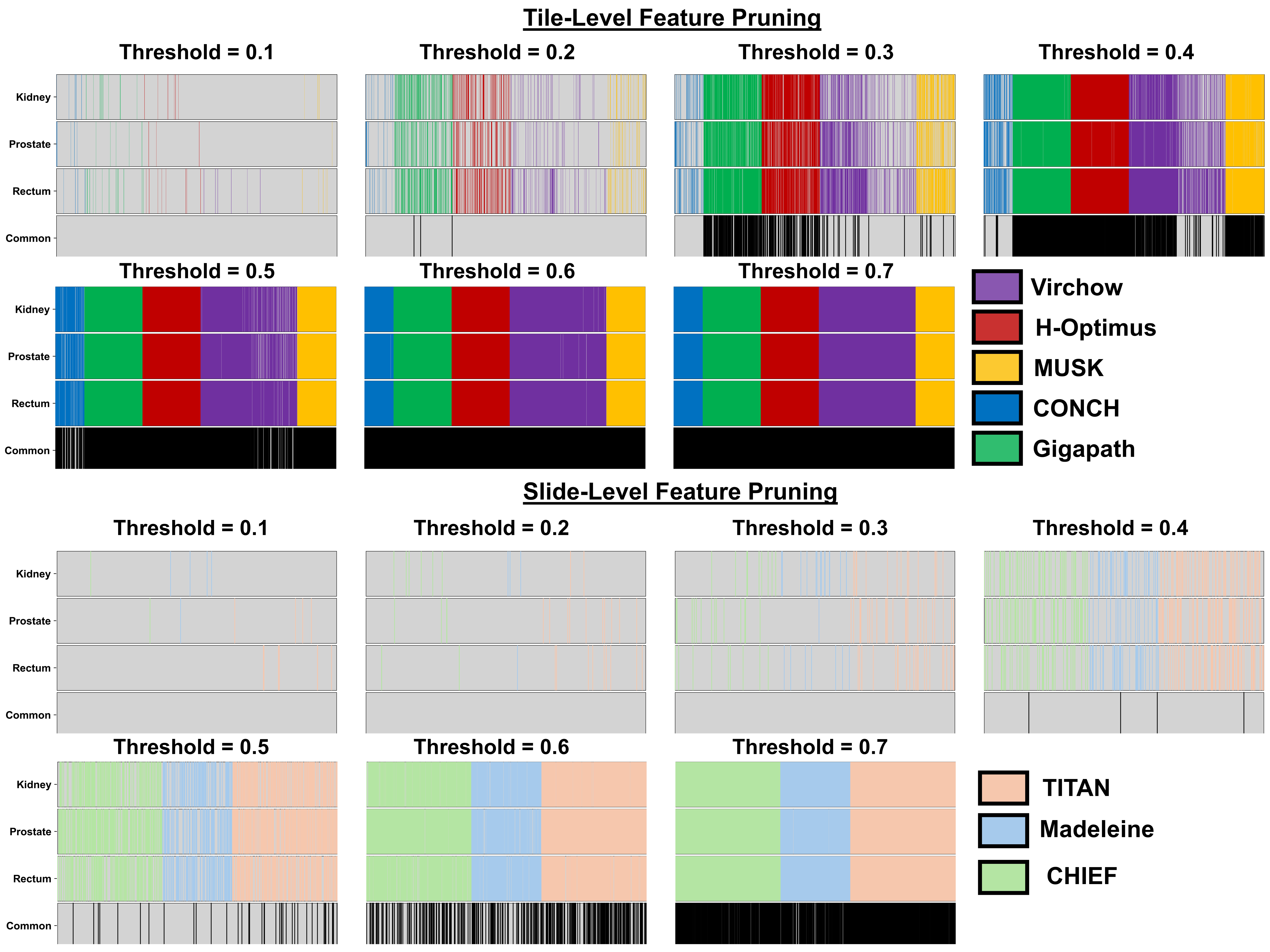} 
\end{center}
\caption 
{ \label{fig:selection}
Visualization of retained foundation model features at different pruning thresholds across kidney, prostate, and rectal cancers for both tile- and slide-level foundation models. Lines represent individual features for different FM models (in different colors), while the black lines in the "common" bar represent features chosen in common across all three disease contexts at that pruning threshold.} 
\end{figure} 

Figure \ref{fig:selection} visualizes commonalities in FM features selected at each pruning threshold, to better understand trends in FM feature redundancy between diseases. No common features are observed between kidney, rectum, and prostate cancers at a tile level at the lowest threshold (0.1), with very few common features are present up to a threshold of 0.3. Prov-gigapath and H-Optimus had the most features commonly chosen across diseases, indicating potentially higher pan-disease relevance. Conch demonstrated the least commonly chosen features across diseases, indicating that Conch features are often correlated regardless of disease context. Much less overlap in chosen features was observed for slide level models (Fig. \ref{fig:selection}). Slide level pruning resulted in almost entirely unique signatures across diseases up to a threshold of 0.4, where TITAN has the most commonly retained features across diseases while MADELEINE had the least features retained. 

\subsection{Experiment 2: Evaluating the similarity of foundation model embeddings}
%To uncover the grounding principals driving the performance improvement of FM fusion observed in Experiment 1, we evaluated the similarity of FM embeddings across diversely trained models. Figure \ref{fig:similarity} portrays average similarity measurements for each FM, gauging the overall similarity of each embedding to other embeddings generated by other FMs. 

%We tested metrics for local similarity (k-NN) which is an indicator of how well small embedding neighborhoods map to each other, and global similarity (CKA, SVCCA, OPD, RR) which indicates how similar groups of embeddings are to eachother. 

\begin{figure}
\begin{center}
\includegraphics[width=\linewidth]{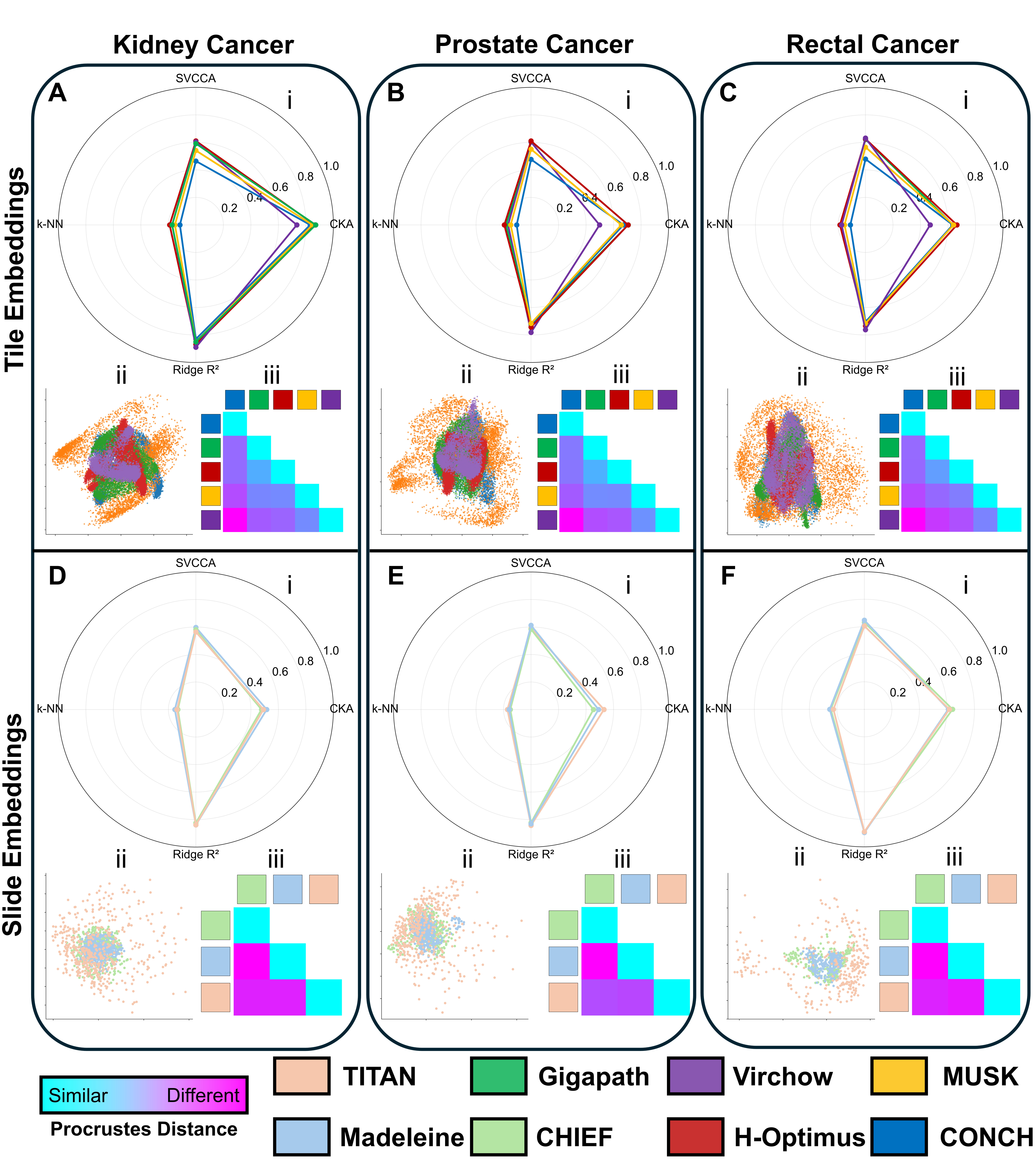}  % fig2 includes two images 
\end{center}
\caption 
{ \label{fig:similarity}
Similarity measurements between foundation model embeddings for tile level models (A-C) and slide level models (D-F). (i) Radar plots visualizing similarity scores as measured by CKA, SVCCA, k-NN, and Ridge $R^2$ (RR). (ii) Scatter plot of FM embeddings (in different colors) overlaid onto one another within the same 2D space. (iii) Cell plot of OPD magnitude where high OPD (purple) indicates dissimilarity and low OPD (blue) indicates similarity between foundation model embeddings.} 
\end{figure}

Global similarity as measured by CKA, RR, SVCCA, and OPD indicate substantial similarity between tile-level FM embeddings for all diseases. 
These can be seen in the CKA scores observed to be greater than 0.6 (with the exception of Virchow), RR scores greater than 0.75, and SVCCA scores greater than 0.4 (Fig. \ref{fig:similarity}A-C, Tab. \ref{tab:tile_similarity}) for all diseases. In contrast, the local similarity of FM embeddings as measured by k-NN scores were observed to be less than 0.2 on average. %CKA, OPD, and k-NN also indicate that H-Optimus and Prov-gigapath produce the most similar tile-level FM representations across all three diseases. Virchow demonstrated the most unique embeddings in terms of CKA, while Conch demonstrated the most unique embeddings in terms of SVCCA and k-NN (Fig. \ref{fig:similarity}A-C). 
Slide level embeddings demonstrated less similarity compared to tile-level embeddings (Fig. \ref{fig:similarity}D-F), with CKA scores of about 0.5, RR scores greater than 0.8, and SVCCA scores of about 0.6. Despite indicating moderate global similarity between model embeddings, models achieved k-NN scores less than 0.2 (corresponding to low local similarity). %Slightly higher similarity was observed in rectal cancer for all metrics (Fig. \ref{fig:similarity} F).%However, both of these scores still indicate mild similarity between model embeddings. 
%The most similar slide-level embeddings were indicated to be CHIEF and MADELEINE by OPD and RR in all diseases, however these trends were mild compared to those seen with tile-level embeddings.
\begin{table*}[t]
\centering
\caption{\label{tab:tile_similarity}Pairwise similarity between tile level-foundation models.}
\setlength{\tabcolsep}{3pt}                % tighten horizontal padding
\renewcommand{\arraystretch}{0.98}         % slightly tighter rows
\begin{adjustbox}{max width=\textwidth,center} % shrink only; never scale up
\small % or \footnotesize if you need more shrink
\begin{tabular}{
l l
S[table-format=1.2]
S[table-format=1.2]
S[table-format=1.2]
S[table-format=1.2]
S[table-format=1.2]
S[table-format=1.2]
}
\toprule
\multicolumn{1}{c}{Model A} & \multicolumn{1}{c}{Model B} &
\multicolumn{1}{c}{CKA} & \multicolumn{1}{c}{SVCCA} &
\multicolumn{1}{c}{Proc.} & \multicolumn{1}{c}{kNN} &
\multicolumn{1}{c}{$R^{2}_{Y\rightarrow X}$} & \multicolumn{1}{c}{$R^{2}_{X\rightarrow Y}$} \\
\midrule

\multicolumn{8}{c}{\textbf{Kidney}} \\
\midrule
Musk     & Virchow  & 0.73 & 0.58 & 0.63 & 0.17 & 0.94 & 0.84 \\
Musk     & Hoptimus & 0.88 & 0.57 & 0.64 & 0.17 & 0.89 & 0.82 \\
Musk     & Gigapath & 0.90 & 0.56 & 0.66 & 0.15 & 0.88 & 0.78 \\
Musk     & Conch    & 0.88 & 0.46 & 0.98 & 0.12 & 0.82 & 0.84 \\
Virchow  & Hoptimus & 0.76 & 0.71 & 0.85 & 0.23 & 0.91 & 0.93 \\
Virchow  & Gigapath & 0.75 & 0.65 & 0.90 & 0.18 & 0.88 & 0.91 \\
Virchow  & Conch    & 0.69 & 0.47 & 1.40 & 0.12 & 0.80 & 0.93 \\
Hoptimus & Gigapath & 0.95 & 0.69 & 0.45 & 0.25 & 0.88 & 0.87 \\
Hoptimus & Conch    & 0.87 & 0.47 & 0.81 & 0.12 & 0.78 & 0.87 \\
Gigapath & Conch    & 0.89 & 0.45 & 0.82 & 0.11 & 0.75 & 0.86 \\

\midrule
\multicolumn{8}{c}{\textbf{Prostate}} \\
\midrule
Musk     & Virchow  & 0.46 & 0.58 & 0.73 & 0.16 & 0.91 & 0.82 \\
Musk     & Hoptimus & 0.77 & 0.56 & 0.78 & 0.17 & 0.84 & 0.75 \\
Musk     & Gigapath & 0.79 & 0.58 & 0.77 & 0.16 & 0.83 & 0.73 \\
Musk     & Conch    & 0.71 & 0.46 & 1.09 & 0.12 & 0.74 & 0.82 \\
Virchow  & Hoptimus & 0.55 & 0.69 & 1.13 & 0.22 & 0.89 & 0.91 \\
Virchow  & Gigapath & 0.49 & 0.65 & 1.18 & 0.18 & 0.87 & 0.89 \\
Virchow  & Conch    & 0.47 & 0.47 & 1.70 & 0.12 & 0.78 & 0.92 \\
Hoptimus & Gigapath & 0.89 & 0.70 & 0.52 & 0.28 & 0.85 & 0.85 \\
Hoptimus & Conch    & 0.68 & 0.47 & 0.92 & 0.13 & 0.71 & 0.87 \\
Gigapath & Conch    & 0.73 & 0.48 & 0.89 & 0.13 & 0.69 & 0.86 \\

\midrule
\multicolumn{8}{c}{\textbf{Rectum}} \\
\midrule
Musk     & Virchow  & 0.61 & 0.61 & 0.66 & 0.18 & 0.93 & 0.81 \\
Musk     & Hoptimus & 0.88 & 0.59 & 0.69 & 0.19 & 0.88 & 0.79 \\
Musk     & Gigapath & 0.86 & 0.59 & 0.79 & 0.17 & 0.87 & 0.72 \\
Musk     & Conch    & 0.89 & 0.47 & 0.97 & 0.14 & 0.80 & 0.83 \\
Virchow  & Hoptimus & 0.61 & 0.73 & 0.93 & 0.24 & 0.88 & 0.92 \\
Virchow  & Gigapath & 0.61 & 0.70 & 1.06 & 0.21 & 0.86 & 0.88 \\
Virchow  & Conch    & 0.60 & 0.48 & 1.36 & 0.13 & 0.76 & 0.92 \\
Hoptimus & Gigapath & 0.92 & 0.72 & 0.52 & 0.27 & 0.87 & 0.83 \\
Hoptimus & Conch    & 0.81 & 0.47 & 0.80 & 0.14 & 0.75 & 0.87 \\
Gigapath & Conch    & 0.80 & 0.47 & 0.80 & 0.14 & 0.66 & 0.86 \\
\bottomrule
\end{tabular}
\end{adjustbox}
\end{table*}

\begin{table*}[t]
\centering
\caption{\label{tab:slide_similarity}Pairwise similarity between slide-level foundation models.}
\setlength{\tabcolsep}{3pt}                % tighten horizontal padding
\renewcommand{\arraystretch}{0.98}         % slightly tighter rows
\begin{adjustbox}{max width=\textwidth,center} % shrink only; never scale up
%\resizebox{\textwidth}{!}{%
\small
\begin{tabular}{
l l
S[table-format=1.2]
S[table-format=1.2]
S[table-format=1.2]
S[table-format=1.2]
S[table-format=1.2]
S[table-format=1.2]
}
\toprule
\multicolumn{1}{c}{Model A} & \multicolumn{1}{c}{Model B} &
\multicolumn{1}{c}{CKA} & \multicolumn{1}{c}{SVCCA} &
\multicolumn{1}{c}{Proc.} & \multicolumn{1}{c}{kNN} &
\multicolumn{1}{c}{$R^{2}_{Y\rightarrow X}$} & \multicolumn{1}{c}{$R^{2}_{X\rightarrow Y}$} \\
\midrule

\multicolumn{8}{c}{\textbf{Kidney}} \\
\midrule
Chief     & Madeleine & 0.50 & 0.61 & 0.94 & 0.15 & 0.85 & 0.80 \\
Chief     & Titan     & 0.45 & 0.55 & 0.81 & 0.12 & 0.92 & 0.73 \\
Madeleine & Titan     & 0.53 & 0.58 & 0.82 & 0.16 & 0.93 & 0.78 \\

\midrule
\multicolumn{8}{c}{\textbf{Prostate}} \\
\midrule
Chief     & Madeleine & 0.41 & 0.61 & 1.10 & 0.14 & 0.80 & 0.82 \\
Chief     & Titan     & 0.49 & 0.56 & 0.77 & 0.16 & 0.90 & 0.76 \\
Madeleine & Titan     & 0.57 & 0.62 & 0.80 & 0.18 & 0.94 & 0.77 \\

\midrule
\multicolumn{8}{c}{\textbf{Rectum}} \\
\midrule
Chief     & Madeleine & 0.63 & 0.67 & 0.91 & 0.27 & 0.90 & 0.89 \\
Chief     & Titan     & 0.65 & 0.59 & 0.78 & 0.22 & 0.97 & 0.80 \\
Madeleine & Titan     & 0.59 & 0.62 & 0.82 & 0.23 & 0.96 & 0.83 \\
\bottomrule
\end{tabular}%
%}
\end{adjustbox}
\end{table*}

\subsection{Experiment 3: Evaluating and interpreting the clinical reasoning of foundation model attention}

%Training models using foundational embeddings from pathology tiles creates two distinct levels of attention: (1) FM encoder attention at the tile level (2) MIL-CLAM attention at the slide level. We set out to evaluate the differences between FM attention at both of these scales to provide mechanical explanations for finding from Experiments 1 and 2. 

\begin{figure}
\begin{center}
\includegraphics[width=\linewidth]{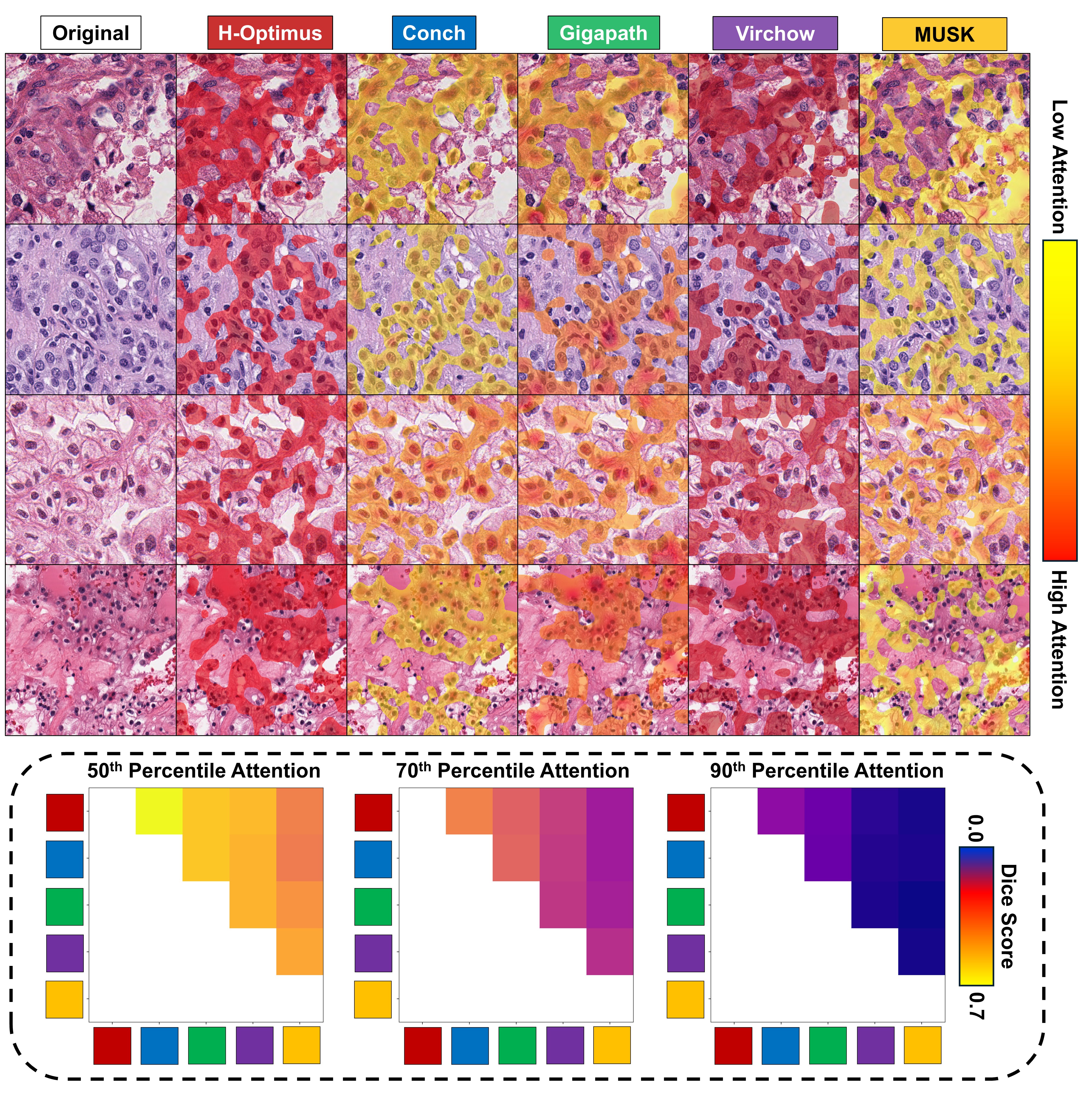}  % fig2 includes two images 
\end{center}
\caption 
{ \label{fig:tile_attention}
Tile-level attention analysis of different FMs. Representative tiles from kidney pathology are shown with attention overlays, where no overlay corresponds to minimal attention, yellow is low attention, and red is high attention. Cell plots visualize similarities in model attentions through measured dice score overlap of binary masks created at different attention cutoffs: 50th, 70th, and 90th percentile. Yellow indicates high overlap while blue indicates low overlap in attention.} 
\end{figure} 

%Findings from Experiment 2 indicate that FM embeddings derived from pathology tiles contain some shared information. 
Figure \ref{fig:tile_attention} visualizes attention maps from each of the tile-level encoders that were tested together with the overlap of their respective attention maps, which demonstrate substantial similarities. For example, H-Optimus and Conch have remarkable visual similarity in attention, while H-Optimus and MUSK exhibit very little similarity. These observations can be confirmed by dice score measurements of attention overlap at different attention thresholds. At the 50th percentile of model attentions, Conch and H-Optimus attention maps achieved the highest dice overlap of 0.65, while Conch and MUSK presented with the lowest overlap of 0.48. When constraining attention to the 90th percentile and above, the maximum overlap was still observed between H-Optimus and Conch (dice = 0.28) while the minimum overlap was observed between MUSK and all other models (dice=0.12-0.13). This indicates that while  FMs attend to some similar areas (50th percentile), but that the most attended regions (90th percentile) by each model are relatively unique. 
 
\begin{figure}[t]
  \centering
  \computeimgmaxheight
  \includegraphics[width=\textwidth,height=21cm,keepaspectratio]{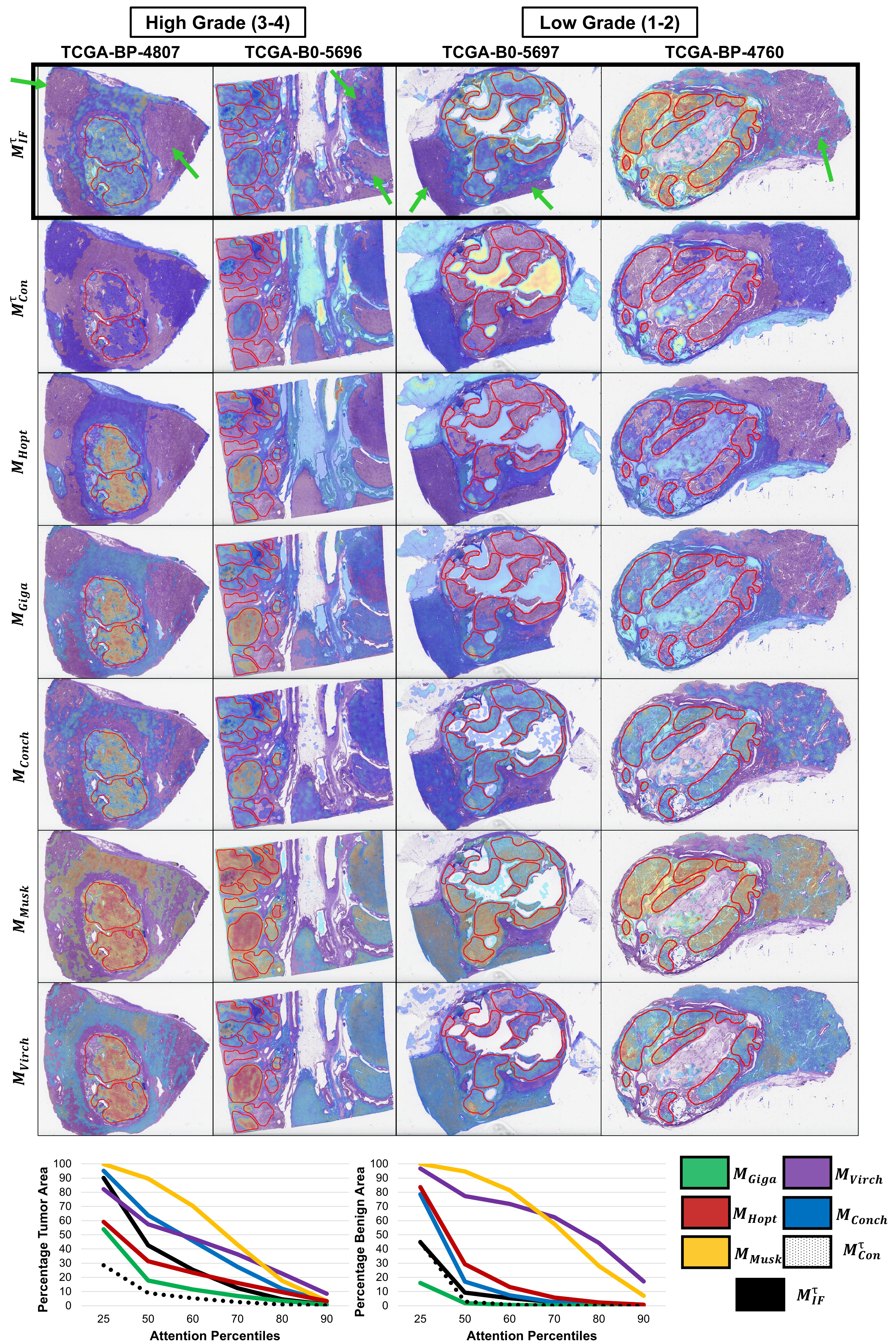}
  \caption{\label{fig:slide_attention}Slide-level attention of individual MIL-CLAM models and fusion strategies. Model attention are shown as a heatmap on a representative kidney pathology image (no overlay: very little attention; blue: low attention; yellow: medium attention; red: high attention). Tumor regions are outlined in red, while green arrows point to known benign regions. Line plots depicts trends in overlap of model attention within tumor and benign regions, across varying attention percentiles. }
\end{figure}

%At the slide level we sought to measure how feature pruning affected slide-level attention from MIL-CLAM models trained on FM features. 
Figure \ref{fig:slide_attention} depicts representative attention maps for high and low grade samples for all individual tile-level models tested as well as integration approaches, in the kidney cancer grade prediction task. Visual observation indicates that intelligent fusion results in $M_{IF}^\tau$ attention that is concentrated on tumor regions while avoiding benign parenchyma. Comparing this to MIL-CLAM models trained on individual FM embeddings, less coverage of tumor regions and persistent attention to clear benign regions can be observed. Some individual models ($M_{Virch}$, $M_{Musk}$) show more intense attention to tumor regions, but still have widespread attention to benign parenchyma, which could be considered a false positive in the context of a grade prediction task. %, which may not have relevance to a grade prediction task. 
$M_{Con}^\tau$ demonstrates poor concentration on tumor tissue, often with spurious attention in regions empty of tissue. These trends are consistent across both high grade and low grade samples. 

%To quantitatively confirm these findings, we calculated the overlap of model attentions at different percentiles with both tumor and benign regions. 
In quantitative evaluation, $M_{IF}^\tau$ demonstrated high attention towards tumor regions (90\% coverage at 25th percentile) while simultaneously avoiding benign regions (44\% coverage at 25th percentile). No other models tested in this study demonstrated high attention to tumor regions while simultaneously avoiding benign regions. Most individual models demonstrated tumor coverage $>$80\% ($M_{Musk}$, $M_{Conch}$, $M_{Virch}$), but all but $M_{Giga}$ also demonstrated spurious attention ($>70\%$) to benign parenchyma. $M_{Con}^\tau$ demonstrated very little attention to both tumor and benign regions, confirming our qualitative observations of spurious attention to empty regions of the slide. 

% and some demonstrated very little tumor coverage ($M_{Hopt}^K$, $M_{Giga}^K$, $M_{Con, T}^K$). $M_{Corr,T}^K$ demonstrate the largest difference in tumor attention moving from 25th percentile to 50th percentile, potentially indicating more concentrated attention on specific high grade tumor regions. Most other models  attended to benign regions at much higher rates than $M_{Corr, T}^K$ ($M_{Musk}^K$: 99\%, $M_{Virch}^K$: 97\%, $M_{Hopt}^K$: 84\%, $M_{Conch}^K$: 79\%) while only $M_{Giga}^K$ had less attention (17\%). $M_{Con, T}^K$ demonstrated similar levels of benign attention compared to $M_{Corr,T}^K$. 

\begin{figure} [ht]
\begin{center}
\begin{tabular}{c} %% tabular useful for creating an array of images 
\includegraphics[width=\linewidth]{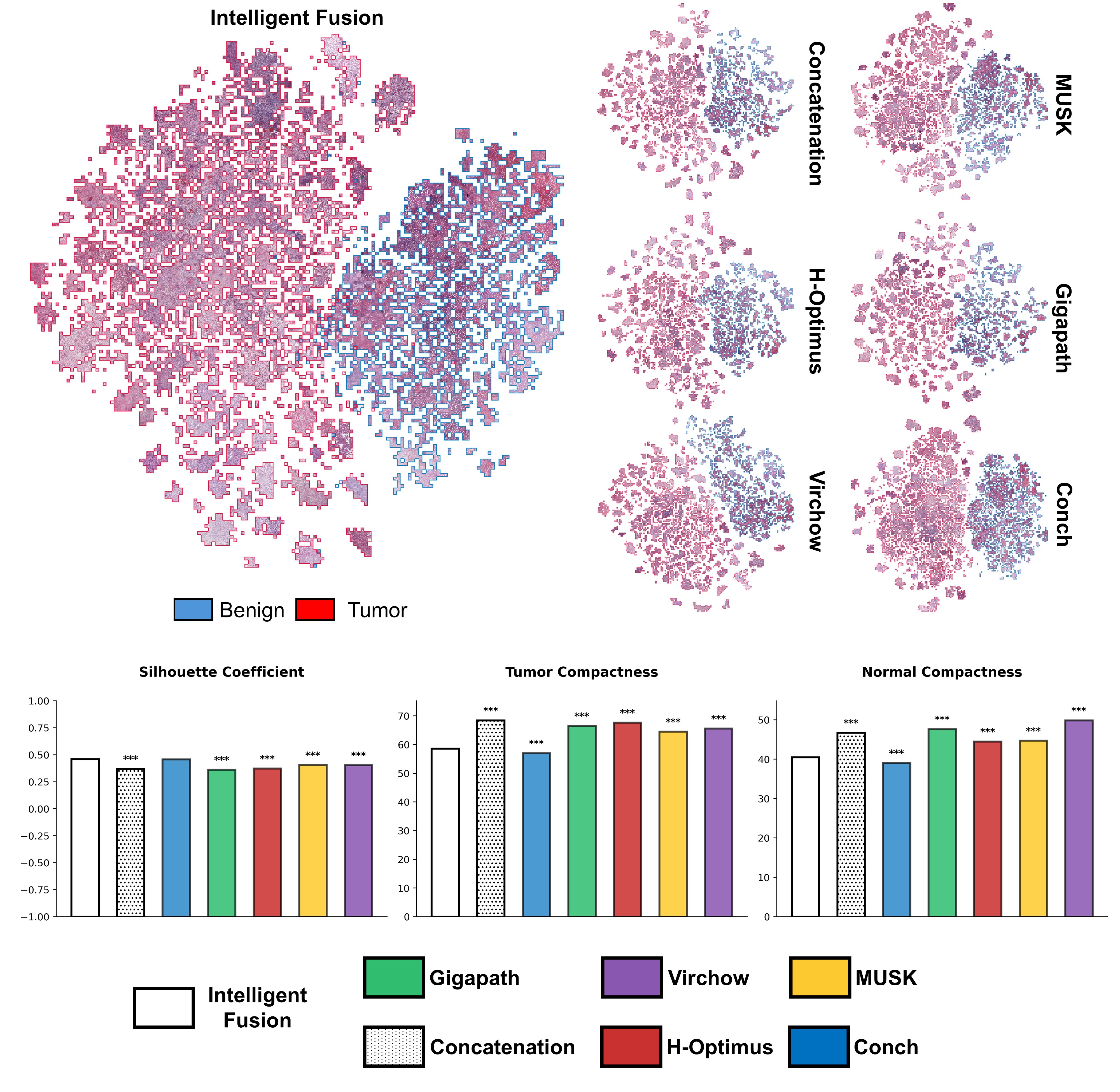}
\end{tabular}
\end{center}
\caption{ 
{ \label{fig:clustering} Qualitative and quantitative clustering analysis of foundation model embeddings. 2D tSNE plots of tiles are shown for all FM embeddings, as well as for naive and intelligent fusion. Bar plots illustrate quantitative analysis of clustering efficacy through silhouette coefficient (higher means improved separation of benign and tumor tissues) as well as benign and tumor clustering compactness (lower indicates more compact clustering). }}
\end{figure} 

Figure \ref{fig:clustering} depicts tile-level clustering results of 2D tSNE embeddings based on different FM signatures. Clusters visually demonstrate that benign and tumor tissues are more clearly separated using the intelligent fusion FM signature compared to other methods. Quantitative analysis further shows that intelligent fusion resulted in significantly improved clustering of benign and tumor tissues via silhouette coefficient (0.48 for IF, $<$ 0.4 for others except Conch), as well as tighter clustering compactness of both tumor (58 for IF, $>$ 65 for others except Conch) and benign tissue (41 for IF, $>$ 45 for others except Conch). %By promoting this separation of tissue types, intelligent fusion could support better attention allocation to vital regions of tumor, avoiding more easily identified benign tissues. 

\section{Discussion}

%_______________________________Tile Level Models___________________________________
The recent development of foundation models in digital pathology have demonstrated significant promise for quantitative characterization of multiple diseases. Toward exploiting the potential complementarity between different FMs, we presented a novel framework for intelligent fusion of pathologic FM representations through systematic pruning and integration of FM embeddings. We showed that our novel fusion approach results in improved characterization of multiple cancers, while directing model attention to specific biologically relevant regions. We provided a basis for gains in information fusion through a comprehensive analysis of  similarity between FM embeddings, demonstrating that diversely trained FMs retain substantially similar signatures. We also showed that redundant signatures are likely sourced from overlapping tissue regions at the tile level as indicated by FM attention. 

Our results demonstrate substantial benefits to fusing FM features or decisions, where fusion techniques provided the best classification performance in all three diseases across both tile- and slide-level models. This is consistent with recent advances in the field, which have reported performance improvements using FM ensembling\cite{zhao_uncertainty-aware_2025} and FM embedding concatenation \cite{neidlinger_benchmarking_2025}. Both of these previous studies demonstrated that FMs capture complementary information that when combined provides a more comprehensive signature of the target tissue. 
However, we specifically found that intelligently fusing FM embeddings by pruning redundant features significantly improves performance over more naive fusion methods as well as individual FM embeddings.  
To our knowledge, our study is the first to examine multiple approaches to interrogating FM embedding redundancy, as well as fusing these reduced embeddings for downstream predictive tasks. %Our findings indicate that correlation pruning of tile level FM embeddings leads to significantly better F1 scores compared to other methods, suggesting that feature pruning encourages balanced performance between classes. This is further confirmed by looking at sensitivity and specificity, on which correlation pruned models consistently outperform other fusion methods and individual FMs. Feature pruning seems to result in either better or similar AUC values compared to other metrics. As a whole, these findings indicate that feature pruning promotes more balanced performance between classes (high and low grade cancers), which is vital for clinical deployment. 

Our findings make intuitive sense when considering the context of how FMs are trained, which involve using substantial quantities of pathology data from diverse diseases and tissues. Their learning objectives are intended to guarantee that FMs will learn features that can distinguish between tissues of different appearance. However, given the sheer magnitude of tissue variability that they are exposed to, it seems unlikely that each feature in an FM embedding would hold relevance for every given disease. It is well known that providing redundant or irrelevant information can detract from model performance \cite{he_clustering-based_2020}, and thus removing this information prior to training is likely to enable better learning for the target task. %Relying on deep learning approaches while additionally sorting through irrelevant information could thus result in sub-optimal downstream models.
This can be observed when comparing the classifier performance of naive concatenation of multiple FMs representations vs intelligent fusion, which suggests representational redundancies may be compounded in the former case. By pruning model embeddings, we posit that our framework enables the capturing of unique sets of information that quantify the underlying pathology while reducing the risk of spurious correlations.

% Slide-level improvements from intelligent fusion can also be seen as a result of reducing redundant information induced by similar training practices and targets. 
% The slightly reduced efficacy of pruning for slide level FMs can specifically be attributed to the lower level of similarity between slide-level embeddings. With lower similarity between embedding spaces, it is more important to retain those different views and fuse them instead of trimming the representations. These lower levels of similarity can potentially be explained by the complexities of aggregating large quantities of information across an entire slide, compounding the dissimilarities present from the tile level embeddings with different aggregation techniques. 

%While some prior studies have shown efficacy from ensembling foundation models\cite{zhao_uncertainty-aware_2025} or fusing their embeddings \cite{neidlinger_benchmarking_2025}, 
While previous works have examined performance differences between FM representations\cite{zhao_uncertainty-aware_2025, neidlinger_benchmarking_2025}, there has not been a comprehensive evaluation of similarity of information content between FM representations and their embedding space. Our experiments revealed that tile level embeddings from different FMs have substantial \textit{global} similarity, as evaluated via multiple embedding similarity measures (CKA, SVCCA, OPD, and RR), indicating clear embedding similarity and alignment between many of the FMs tested.
%These metrics measure \emph{global} similarity, which measures the alignment of the embedding representations between two models across the entire dataset. 
However, when one looks at \emph{local} similarity (via the k-NN measure), substantially less similarity can be observed. Local similarity measures indicate how well FM embeddings align within specific small neighborhoods of tiles, and our results suggest that FMs harbor unique local signatures which could drive subtle differences between them. % and therefore downstream performance. 
This illustrates that while FMs do indeed pick up similar signatures of tissue, each model additionally includes some unique information not held within other model embeddings. 

Critically, there has been only limited study of the specific drivers of performance when ensembling FMs. Neidlinger et al\cite{neidlinger_benchmarking_2025} demonstrated that MIL models trained using different tile level FM features attended to different regions of the slide, and that their downstream prediction scores prior to the final model activation function held mild to medium levels of similarity. Expanding on this, our more comprehensive experiments suggest that these embedding differences may originate from the specific regions attended by FMs at the tile level. We found that FMs attend to some similar regions on tiles but also have distinct differences, indicated by overlaps in the 50th percentile of tile-level attention (45-55 \% dice). %which aligns with the similarity metrics we observed in Experiment 2. 
We also observed that this dice overlap diminished dramatically to 12-28\% at the 90th percentile of attention, indicating that while FMs show some similarities in attention maps, specific regions of importance are different between them. 
%This differential weighting of the same regions on the tile level is a potential explanation for the level of similarity we observed in FM embeddings. 
%While prior studies have established that fusing FMs can improve performance, it is unclear how fusing FM embeddings alters the attention of the downstream MIL models. 
Our findings also demonstrated that intelligent FM fusion via correlation pruning aids in focusing MIL model attention towards tumor regions while reducing spurious attention towards benign regions for cancer grading. %All individual models demonstrated clear attention towards benign parenchyma despite attempting to predict cancer grade. %Some models miss tumor regions entirely and attend to empty regions without tissue. 
Concatenation led to attention being dispersed into benign regions with very little attention towards the target tumor regions, indicating that simple combination may divert attention away from the desired regions for the target problem. %Intelligent fusion demonstrated remarkable ability to avoid  benign regions while also maximally attending to tumor regions. This result was both qualitatively and quantitatively confirmed. 
This indicates that correlation pruning not only improves performance, but that it concentrates model attention towards interpretable and desirable regions of tissue in ways that simple concatenation does not achieve. Intelligent fusion of signatures also enhanced unsupervised clustering of benign and tumor tissue compartments across the dataset, yielding more separable representations. By refining the feature space to better distinguish target tissue types prior to training, this fusion likely primes the model to allocate more focused attention at the slide level.

Empirically, we found that the performance of our fusion technique was impacted by the pruning threshold, which can be treated as an optimizable hyperparameter for FM fusion. Studying trends in FM representation pruning also revealed information about specific FM embeddings. Conch was more consistently retained compared to other models at the relaxed threshold of 0.1, but was then pruned substantially at higher thresholds, indicating that the Conch vector may comprise only a small subset of important and uncorrelated features. For other models, such as Virchow and MUSK, the number of retained features increased smoothly as the threshold became more permissive, suggesting that correlated features are distributed relatively evenly across their embeddings rather than being concentrated in a small subset. Our findings also revealed that these correlations depend on disease context. %Comparing diseases, it is apparent that different percentages of features are retained at each threshold, indicating that feature correlations potentially shift when those features are extracted from different tissues. Slide level models presented with much less consistent feature retention across diseases, potentially indicating that slide level features can change dramatically based on disease context. %These findings are further supported by Figure \ref{fig:selection}, which clearly demonstrates that different features are chosen at each threshold across diseases. This illustrates that different FM features hold significance for different diseases, and that correlation pruning could potentially orient an FM embedding towards a specific tissue type. 
Comparing the rates at which features are pruned in slide models to tile models, we see that slide model features are heavily pruned ($<$ 10\% of features) up to thresholds of 0.4, unlike tile level models which reach 50\% of features at 0.4. %This indicates that slide-level features may be more significantly correlated, which at first glance contradicts our findings of reduced similarity. 
This is likely because slide level FM features comprise two subsets of features: (1) a large set of highly redundant features, and (2) a smaller set of very significant, non-redundant features. The trade-off in pruning these two sets of features (as well as the specific features retained) appears to change based on disease context. %The first set are highly pruned, while the second set differentiates model signals. We know that these two sets of features change based on disease context for slide models, but selected slide features are substantially different for each disease up to much higher thresholds compared to tile models. 

We do note some limitations to our study. First, this study includes only eight of the many currently publicly released FMs. %While a more extensive comparison of FM fusion using these various models is warranted, adding more complexity to the comparisons completed in this study would make drawing overarching conclusions difficult. 
This choice was made for the sake of implementation complexity. We decided to limit the study to a portion of the most prominent models in the field. Second, this study suggests that there may be other possible combination mechanisms, including combining models across scales (slide + tile level) that are beyond the scope of the current study but are nonetheless important targets of inquiry. Future studies will explore different and more complex fusion strategies, validate them in additional holdout datasets, and link model decision making to more specific aspects of underlying biology. 

\section{Concluding Remarks}
In this study, we presented a novel approach to intelligent fusion of pathological foundation models to facilitate improved model performance across multiple diseases, conditions, and scales. Through a detailed interrogation of information content of different FMs, we confirmed that foundation model embeddings contain similar information, and that information is derived from similar tile-level attention. Our findings suggest that our intelligent fusion approach works by minimizing redundant common information between embeddings while retaining critical signatures associated with specific diseases and tissues. 
We further demonstrated that this approach not only drives performance improvements but instills distinct interpretability in model attention. %Feature pruning significantly outperforms individual foundation models and simple fusion mechanisms on most tasks. TWe establish that pruning isolates disease specific signatures from different foundation models, facilitating improved downstream performance on specific tasks. 
Future studies will explore more complex mechanisms of foundation model fusion to improve downstream disease characterization, as well as validate our findings in other disease contexts and additional holdout datasets. 

\section{Acknowledgments}

Research reported in this publication was supported by the National Cancer Institute (1R01CA280981-01A1, 1U01CA294415-01A1, 1F31CA291057-01A1), the National Institute of Nursing Research (1R01NR019585-01A1), the National Institute of Biomedical Imaging and Bioengineering (T32EB007509, 1R01EB037526-01), the National Heart, Lung, and Blood Institute (1R01HL165218-01A1), the National Science Foundation (Award 2320952), the Veterans Affairs Biomedical Laboratory Research and Development Service (1I01BX006439-01), the DOD Peer Reviewed Cancer Research Program (W81XWH-21-1-0725), the Leona M. and Harry B. Helmsley Charitable Trust, the Ohio Third Frontier Technology Validation Fund, the JobsOhio Program, and the Wallace H. Coulter Foundation Program in the Department of Biomedical Engineering at Case Western Reserve University. This work made use of the High Performance Computing Resource in the Core Facility for Advanced Research Computing at Case Western Reserve University.

The content is solely the responsibility of the authors and does not necessarily represent the official views of the National Institutes of Health, the U.S. Department of Veterans Affairs, the Department of Defense, or the United States Government.

\bibliography{fm_analysis}
\bibliographystyle{splncs03_unsrt}

\vspace{12pt}

\end{document}